\documentclass[final,3p,times]{elsarticle}

\usepackage{amssymb}
\usepackage{amsmath,times,bm}
\usepackage{comment}
\usepackage[subrefformat=parens]{subcaption}
\usepackage{here}

\usepackage{lineno}
\usepackage{multirow}
\usepackage{xcolor}

\journal{Engineering with Computers}

\begin{document}

\begin{frontmatter}



\title{A finite element-based physics-informed operator learning framework for spatiotemporal partial differential equations on arbitrary domains}


\author[label1]{Yusuke Yamazaki\corref{cor1}}
\author[label2]{Ali Harandi}
\author[label3]{Mayu Muramatsu}
\author[label4]{Alexandre Viardin}
\author[label4]{Markus Apel}
\author[label2]{Tim Brepols}
\author[label2]{\\Stefanie Reese}
\author[label4]{Shahed Rezaei\corref{cor1}} 

\affiliation[label1]{organization={Graduate School of Science and Technology, Keio University},
            addressline={Hiyoshi3-14-1}, 
            city={Kohoku-ku, Yokohama},
            postcode={223-8522}, 
            country={JAPAN}}
\affiliation[label2]{organization={Institute of Applied Mechanics, RWTH Aachen University},
            addressline={Mies-van-der-Rohe-Str. 1}, 
            city={Aachen},
            postcode={52074}, 
            country={GERMANY}}
\affiliation[label3]{organization={Department of Mechanical Engineering, Keio University},
            addressline={Hiyoshi3-14-1}, 
            city={Kohoku-ku, Yokohama},
            postcode={223-8522}, 
            country={JAPAN}}
\affiliation[label4]{organization={ACCESS e.V.},
            addressline={Intzestr. 5}, 
            city={Aachen},
            postcode={52072}, 
            country={GERMANY}}
\cortext[cor1]{yusuke.yamazaki.0615@keio.jp, s.rezaei@access-technology.de}    

\begin{abstract}
We propose a novel finite element-based physics-informed operator learning framework that allows for predicting spatiotemporal dynamics governed by partial differential equations (PDEs). The Galerkin discretized weak formulation is employed to incorporate physics into the loss function, termed finite operator learning (FOL), along with the implicit Euler time integration scheme for temporal discretization. A transient thermal conduction problem is considered to benchmark the performance, where FOL takes a temperature field at the current time step as input and predicts a temperature field at the next time step. Upon training, the network successfully predicts the temperature evolution over time for any initial temperature field at high accuracy compared to the solution by the finite element method (FEM) even with a heterogeneous thermal conductivity and arbitrary geometry. The advantages of FOL can be summarized as follows: First, the training is performed in an unsupervised manner, avoiding the need for large data prepared from costly simulations or experiments. Instead, random temperature patterns generated by the Gaussian random process and the Fourier series, combined with constant temperature fields, are used as training data to cover possible temperature cases. Additionally, shape functions and backward difference approximation are exploited for the domain discretization, resulting in a purely algebraic equation. This enhances training efficiency, as one avoids time-consuming automatic differentiation in optimizing weights and biases while accepting possible discretization errors. Finally, thanks to the interpolation power of FEM, any arbitrary geometry with heterogeneous microstructure can be handled with FOL, which is crucial to addressing various engineering application scenarios.
\end{abstract}




\end{frontmatter}
\section{Introduction}
Over the past decade, machine learning (ML) methods have played a prominent role in scientific and engineering applications. They can learn how to perform a given task that previously could only be done by humans. The famous applications include self-driving cars \cite{gupta2021deep}, natural language processing \cite{collobert2011natural}, and image recognition \cite{krizhevsky2012imagenet}. Many of these applications utilize deep learning (DL), a subset of ML that has attracted attention for its ability to perform the mapping between input and output features. 
Accurate prediction by DL can be achieved by sufficiently training artificial neural networks (NNs).

When it comes to the field of computational mechanics, there are various types of boundary value problems (BVP) described by partial differential equations (PDEs), which are commonly solved using numerical methods such as the finite element method (FEM), finite difference method, mesh-free method, etc. The problems with the use of numerical schemes can be the computational cost, the complexity of the mesh generation, and, more importantly, the fact that each simulation has to be performed almost from scratch for every scenario \cite{Karniadakis.2021}. DL has been utilized as a promising alternative to avoid these problems. Two mainstreams exist in the domain of DL applications to computational mechanical problems. One is supervised learning based on available labeled data. 
For example, Liang et al. demonstrated the potential of a DL model to be a surrogate of FE analysis in estimating the stress distribution of cardiovascular vessels, which allows for fast and accurate predictions of stress distribution in biomedical applications~\cite{liang2018deep}. 
DL for biomedical applications has also been seen in the prediction of adolescent idiopathic scoliosis \cite{tajdari2021image} and pediatric spinal deformities \cite{tajdari2022next}, in which X-ray data is used as clinical input and the results of calculations by FEM are employed as mechanistic input. One can also see the employment of graph neural networks for the prediction of material concentration in neurite networks \cite{li2021deep} and the combination of the isogeometric analysis with convolutional neural networks (CNNs) for the prediction of neuron growth \cite{qian2023biomimetic}.  In addition, Li et al. proposed an encoder-decoder-based CNN for reaction-diffusion systems as a fast and accurate surrogate tool to FEM \cite{li2020reaction}. 
In material modeling, Hsu et al. presented a DL-based predictive model for crack propagation of crystalline materials using image data processed from the visualized results of molecular dynamics (MD) simulations \cite{hsu2020using}. Furthermore, Fernandez et al. proposed a DL model on the constitutive behavior of grain boundaries, which takes the traction-separation effects into account, based on the data obtained from MD simulations \cite{fernandez2020application}.
Studies have also been conducted that delve into learning differential operators for PDEs from data~\cite{bar2019learning,prakash2024data}. 
Many other DL models on computational mechanical problems have also been developed within the scope of supervised learning; see \cite{bhaduri2022stress,mianroodi2021teaching,mianroodi2022lossless,wang2018multiscale,linka2021constitutive, hagen2023cann} as examples. 

The other common approach is unsupervised learning based on governing equations of BVPs, even without labeled data for training. Originally proposed by Raissi et al., NNs trained based on physics-based loss functions from BVPs are called physics-informed neural networks (PINNs) \cite{Raissi.2019}. 
The key idea is to incorporate governing PDEs directly into the loss functions of NNs with the power of automatic differentiation. Upon successful training, PINNs can accurately predict physical behaviors within the domain of a problem. The training can be seen as the minimization problem in which the residual of PDEs is used as a target function. For the last five years, many researchers have tested the capability of PINNs to predict the behavior of a physical system. Jin et al. developed a PINN framework for incompressible Navier-Stokes equations and verified its capability of obtaining approximate solutions to ill-posed problems with noisy boundary conditions and inverse problems in the context of flow simulation \cite{jin2021nsfnets}. Mao et al. modeled high-speed flow based on the Euler equation using PINNs \cite{mao2020physics}. Mahmoudabadbozchelou et al. presented non-Newtonian PINNs for solving coupled PDEs for fluid while considering the constitutive relationships \cite{mahmoudabadbozchelou2022nn}. Besides, many other studies on fluid-oriented applications of PINNs, such as \cite{rao2020physics,almajid2022prediction,cheng2021deep,eivazi2022physics}, have already been investigated in recent years. 
For heat conduction problems, Zobeiry et al. applied the PINN architecture to the heat transfer equation with convective boundary conditions \cite{Zobeiry.2021}. Cai et al. modeled heat convections with unknown boundary conditions and the two-phase Stefan problem \cite{cai2021physics}. Zhao et al. developed a combined framework of PINNs and CNNs for predictions of temperatures from the information of heat source \cite{zhao2023physics}. Furthermore, Guo et al. worked on the prediction of three-dimensional transient heat conduction targeted for functionally graded materials using the deep collocation method for space and the Runge-Kutta scheme for time integration, showing the applicability of PINN approaches to spatiotemporal three-dimensional complex geometry cases  \cite{guo2023physics}. Readers can also refer to \cite{liu2022temperature,oommen2022solving,he2021physics,manavi2023enhanced,billah2023physics} for other PINNs examples on heat transfer problems. 
When it comes to solid mechanics problems, Samaniego et al. developed a variational energy-based physics-informed loss function for the classical linear elasticity problem and the phase-field model for fracture \cite{Samaniego.2020}. Abueidda et al. used the collocation method to solve solid mechanics problems with various types of material models, including hyperelasticity with large deformation \cite{abueidda2021meshless}. Haghighat et al. demonstrated the applicability of PINNs to the von Mises plasticity model in their PINN framework \cite{Haghighat.2021}. Rezaei et al. proposed a PINN solver for solid problems with heterogeneous elasticities \cite{Rezaei.2022}. 
Harandi et al. solved the thermomechanical coupled system of equations in the heterogeneous domains \cite{Harandi.2023}. Bai et al. developed a modified loss function using the least squares weighted residual method for two- and three-dimensional solid mechanics, which can predict well the displacement and stress fields \cite{bai2023physics}. Other investigations into PINNs for solid mechanics can also be found in~\cite{zhang2022analyses,diao2023solving}. 
In addition, the idea of PINNs has also been combined with the isogeometric analysis for predicting material transports in neurons \cite{li2023isogeometric}.

While previous works on PINNs have provided many discoveries and insights, it is vital to address some drawbacks to enhance applications. For example, a review paper pointed out that PINNs could fail to learn complex physics such as turbulent or chaotic phenomena \cite{Cuomo.2022}.
Wang et al. provided a theoretical analysis of the convergence rate of loss terms in PINNs, revealing the reason why training PINNs may fail in some problem setups \cite{Wang.2022b}. They proposed a neural tangent kernel-based loss-balancing method that reduces the effects of convergence rate discrepancies. Furthermore, PINNs need to be retrained when one wants to consider different boundary conditions or problem domains, although transfer learning can be utilized in this context \cite{Rezaei.2022, xu2023transfer, tang2022transfer}.
As a new DL model that can avoid the latter problem, operator learning has been investigated in recent years as a surrogate for PDE solvers \cite{Lu.2021,Wang.2022,kovachki2023neural,boulle2023mathematical}. The idea is to learn an operator that maps between infinite dimensional Banach spaces. Examples are Fourier neural operators  (FNO) \cite{li2021physics,rashid2022learning}, deep Green networks \cite{gin2021deepgreen,boulle2022data} and deep operator networks (DeepONets) \cite{Goswami.2022,He.2023b,yin2022simulating}. Operator learning can be done in both supervised and unsupervised manners. In the latter case, Wang et al. introduced a physics-informed DeepONet in which physics-informed loss functions from PDEs are used to train the neural operators \cite{Wang.2021b}. Koric et al. compare the performance of the data-driven and physics-informed DeepONets for the heat conduction problem with parametric source terms \cite{koric2023data}. Li et al. introduced a physics-informed version of FNO that works in a hybrid manner to leverage known physics in FNO \cite{li2021physics}. 

Another open problem in physics-informed deep learning is the failure to predict time-dependent evolutionary processes. Wang et al. argued that the causality of physics must be respected in training PINNs when one considers time-continuous problems~\cite{wang2024respecting}. This is the case when we directly treat the temporal dimension as an additional dimension to the spatial domain. Mattey et al. developed a PINN model that enforces backward compatibility over the temporal domain in the loss function to overcome this limitation in the Allen-Cahn and Cahn-Hilliard equations~\cite{Mattey.2022}. 
Xu et al. utilized transfer learning for DeepONet to train the networks with better stability than the original DeepONet for dynamic systems \cite{xu2023transfer}. 
Li et al. presented a phase-field DeepONet, which aims to predict the dynamic behavior of phase-fields in the Allen-Cahn and Cahn-Hilliard equations using the concept of gradient flows \cite{Li.2023}. 
In the latter framework, the trained networks work as an explicit time-stepper that can predict the evolution of the phase field at the next step based on the current phase field. Furthermore, an emerging approach for spatiotemporal predictions is the utilization of numerical discretizations or convolutions to discretize derivatives to learn a discrete mapping on a discretized domain. This direction can also enhance training efficiency by avoiding time-consuming automatic differentiation, especially when higher-order derivatives need to be computed. 
For static problems, Fuhg et al. proposed a deep convolutional Ritz method as a surrogate of numerical solvers, in which the convolution is exploited to take central differences, and the network takes the energy form as a physics-informed loss~\cite{Fuhg.2023}. Gao et al. utilized CNN architecture to deal with the discretized domain and extended it to irregular domains through coordinate transformation~\cite{gao2021phygeonet}. Rezaei et al. devised a framework that they named finite operator learning (FOL) based on FEM for parametrically solving PDEs with a demonstration for a steady heat equation with heterogeneity \cite{rezaei2024integration}. Some other works have applied FEM to integrate the weak-form loss into NNs, such as for advection-diffusion \cite{mitusch2021hybrid}, quantification of wind effects on vibrations \cite{meethal2023finite}, etc. Furthermore, Khara et al. employed the energy-form loss in the FEM-inspired loss function and demonstrated its performance in Poisson's equation including a three-dimensional case \cite{khara2024neufenet}.
When it comes to spatiotemporal problems, Geneva et al. proposed a CNN-based framework with autoregressive encoder-decoder architecture, whose performance is showcased for some types of dynamic PDEs \cite{geneva2020modeling}. Ren et al. presented a discrete learning architecture that combines CNN with long short-term memory for spatiotemporal PDEs \cite{ren2022phycrnet}. Liu et al. embedded known PDE information into CNN architecture itself to preserve the behavior of the PDE of interest for spatiotemporal dynamic phenomena \cite{liu2024multi}. Furthermore, Xiang et al. employed graph neural networks in combination with radial basis function finite difference to predict spatiotemporal dynamics for irregular domains \cite{xiang2024solving}.
The abovementioned works have shown the capability of discrete mapping learned by NNs for spatiotemporal dynamics. However, the researchers in this domain are still looking for approaches that can easily address irregular domains, as it is difficult for CNN-based methods in particular. In this sense, the direction of the incorporation of FEM into discrete operator learning for parametrically solving spatiotemporal PDEs is beneficial to address more realistic problem setups. 
\begin{figure}[t]
    \centering
    \includegraphics[width=160mm]{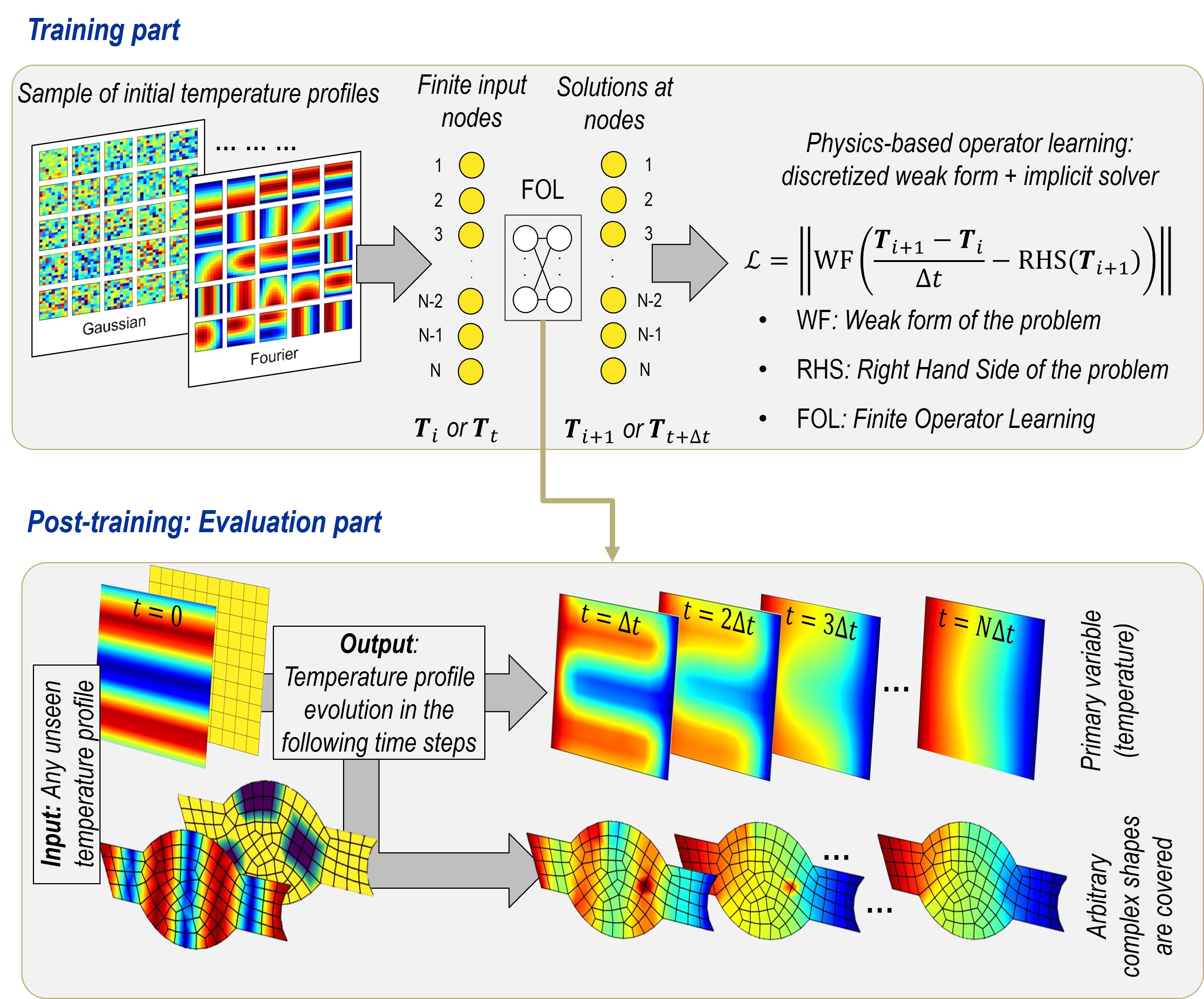}
    \caption{Schematic of training and evaluation parts in the proposed finite element-based physics-informed operator learning framework termed finite operator learning (FOL).  }
    \label{fig:schematics_data-physics}
\end{figure}

In this study, we aim to develop a novel physics-informed discrete-type operator learning framework, which we refer to as FOL, that can parametrically predict the dynamic behavior of physical quantities over time. The schematic illustration of the developed framework is provided in Fig.~\ref{fig:schematics_data-physics}. The key idea is to provide physical fields of a system at the current time step as input and return those at the next time step as output, realizing a surrogate model of time-marching numerical schemes. The time-dependent heat equation, also known as a transient heat conduction problem, is chosen as the target BVP to validate the framework proposed in this work. Not only does this study consider homogeneous thermal conductivity, but it also takes into account heterogeneous conductivity. The training follows a physics-informed loss function constructed based on the finite element discretization of the heat equation \cite{rezaei2024integration}, thereby making it unsupervised learning without labeled data. The extension of the framework to irregular domains is also tested at the end. 

\begin{figure}[H]
    \centering
    \includegraphics[width=140mm]{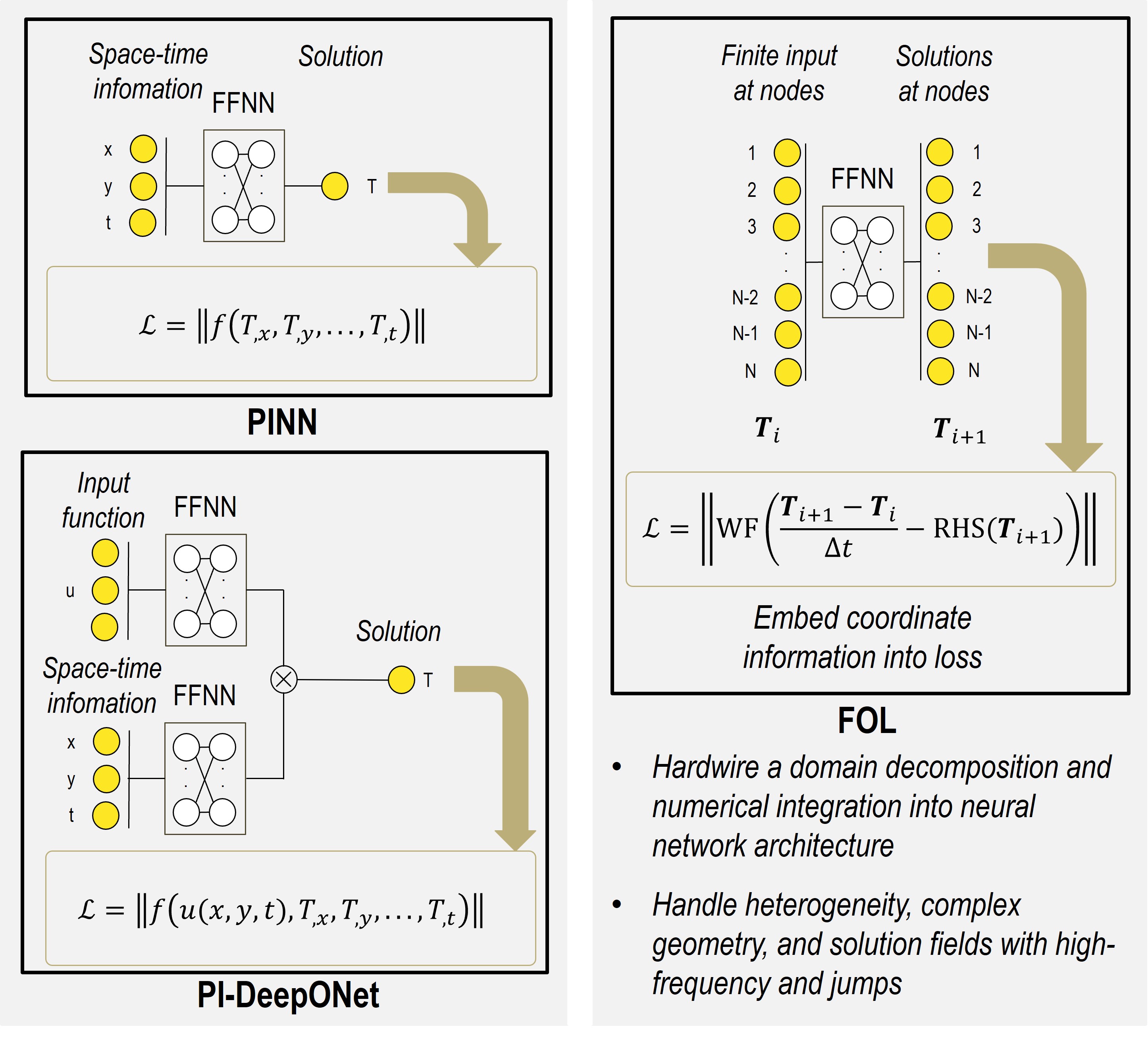}
    \caption{Comparison of vanilla PINNs, physics-informed DeepONet (PI-DeepONet), and finite operator learning (FOL).}
    \label{fig:schematics_comparison}
\end{figure}

The difference from the previous frameworks, such as the one by \cite{Li.2023} or by \cite{liu2024multi} for example, is that this framework directly uses the discretized weak form loss that is identical to the formulation when solving with FEM. This is also demonstrated in a representative model later in this paper. Furthermore, this study considers the heterogeneity of physical properties, which is not addressed in the aforementioned studies.
For clarity, the comparison of the architecture with the vanilla PINNs and physics-informed DeepONet (PI-DeepONet) is described in Fig.~\ref{fig:schematics_comparison}. The pivotal difference is that in FOL we embed the coordinate information into the loss function, allowing us to integrate the branch and trunk nets in DeepONet into a single network. In addition, we do not take the temporal coordinate as input unlike PINNs or PI-DeepOnet; FOL takes into account the temporal evolution by discretizing a given PDE in time between the current and next time steps.

This paper consists of five sections. Section 1 describes the background in the field of scientific machine learning with a focus on physics-informed deep learning and the objective of the present work. Section 2 briefly summarizes the formulation of the discretized heat equation in a weak form by FEM. Following that, the methodology, including the problem setup, developed operator learning framework, and procedure of the training data generation, is explained in Section 3. The results and discussion on the performance of the present framework, as compared to the reference solution by FEM, are reported in Section 4. Finally, the conclusion of the present work is provided along with the outlook in Section 5.
\section{Discretized weak form of heat equation}
In this work, we consider the transient heat conduction problem, which is described by the heat equation, as a benchmark problem to demonstrate the ability of the proposed framework. 
The heat equation describes how the temperature $T(\bm x, t)$, with $\bm x$ being the position and $t$ the time, evolves in the domain  $\bm x \in \Omega$ over time. 
Let the heat source be $Q: \Omega \times \ \left(0,~\tau\right) \rightarrow \mathbb R$, 
the boundary temperature $T_d(\bm x): \ \Gamma_d \ \times  \ \left(0,\tau\right) \rightarrow \mathbb R$, and the boundary heat source $q_n: \Gamma_n \times \ \left(0,\tau\right) \rightarrow \mathbb R$, where $\Gamma_d$ is the domain on which the Dirichlet boundary condition is applied, $\Gamma_n$ is the domain on which the Neumann boundary condition is applied, and $t \in \left(0,\tau\right)$ denotes the open range of the temporal domain with $\tau$ the end. The strong form of the heat equation is given as,
\begin{equation}
c \rho  \dot{T}(\bm x, t)= -\text{div}(\bm q) + Q \quad \text{in} \ \Omega \ \times  \ \left(0,\tau\right),
\end{equation}
where $c$ is the specific heat capacity, $\rho$ is the density, $\dot{T}$ represents the first-order partial derivative with respect to time, and $\bm q = -{k(\bm x)} \nabla T(\bm x,t)$ is the heat flux with ${k(\bm x)}$ the position-dependent thermal conductivity. 
The boundary and initial conditions are enforced by,
\begin{equation}
    T(\bm x,t) = T_d(\bm x)  \quad \text{on} \ \Gamma_d \ \times  \ \left(0,\tau\right),
\end{equation}
\begin{equation}
    \nabla T(\bm x,t) \cdot \bm n = q_n(\bm x)   \quad \text{on} \ \Gamma_n \ \times  \ \left(0,\tau\right),
\end{equation}
\begin{equation}
    T(\bm x,0) = T_0(\bm x) \quad \bm x \in \Omega,
\end{equation}
where $\bm n$ is the outward normal vector.
After multiplication by a test function, taking the integral over the domain and applying Gauss theorem, and assuming no heat source term $Q$ and heat influx and outflux $q_n$, one can obtain the corresponding weak form for $\Omega \ \times  \ \left[0,~\tau\right]$ as,
\begin{equation}
    \int_{\Omega} w c \rho \dot T  d V + \int_{\Omega} {\nabla w}^T {k(\bm x)} \nabla T d V = 0,
\end{equation}
with the initial condition
\begin{equation}
    \int_{\Omega} w c \rho T(t=0) d V= \int_{\Omega} w c \rho T_0  d V,
\end{equation}
where $w$ is the test function defined on an appropriate function space. 
With the weak form at hand, one can arrive at the discretized weak form by the finite element method as, 
\begin{equation}
    \left ( \bm M + \alpha {\Delta t} \bm K \right) \bm T^{n+1} = \left(\bm M - (1-\alpha)\Delta t \bm K \right)\bm T^{n},
\label{eq:FE_heat}
\end{equation}
where 
\begin{equation}
    \bm M = \int_{\Omega} \bm N^T (\rho c) \bm N dV,
\end{equation}
\begin{equation}
    \bm K = \int_{\Omega} \bm B^T k (\bm x) \bm B dV.
\end{equation}
In the formulation above, $\alpha$ is the parameter that can be selected from $0, 0.5, 1$ depending on the choice of time integration scheme, and $\bm T$ is the vector storing nodal temperature values, and the superscript $n$ is used to denote the number of time step increments. Here we introduce the shape function $\bm N$, its spatial derivative $\bm B$, and thermal conductivity $k(\bm x)$. At the element level, they are defined in the case of iso-parametric quadrilateral elements as,
\begin{equation}
    \bm N_e = \left[N_1 \cdots N_4 \right],
\end{equation}
\begin{equation}
 \bm{B}_e=\left[\begin{array}{lll}
N_{1, x} & \cdots & N_{4, x} \\
N_{1, y} & \cdots & N_{4, y}
\end{array}\right],
\end{equation}
\begin{equation}
    \bm k_e = \left[k_1 \cdots k_4 \right]^T,
\end{equation}
where $N_i$ and $k_i$ denote the shape function and thermal conductivity for node $i$ and the subscript $e$ represents the element number. The thermal conductivity is interpolated for Gaussian integration by the nodal thermal conductivity values using the shape function
\begin{equation}
    k(\bm x) = \bm N_e \bm k_e.
\end{equation}
It is worth noting that in the practical implementation, one has to manipulate the left-hand side matrix $\left ( \bm M + \alpha {\Delta t} \bm K \right)$ and the right-hand side $\left(\bm M - (1-\alpha)\Delta t \bm K \right)\bm T^{n}$ based on the Dirichlet boundary conditions to appropriately impose fixed temperatures at desired boundary nodes.   
\section{Methodology}
\subsection{Problem setup}
\begin{figure}[b]
  \begin{minipage}{0.49\textwidth}
    \centering
    \includegraphics[width=74mm]{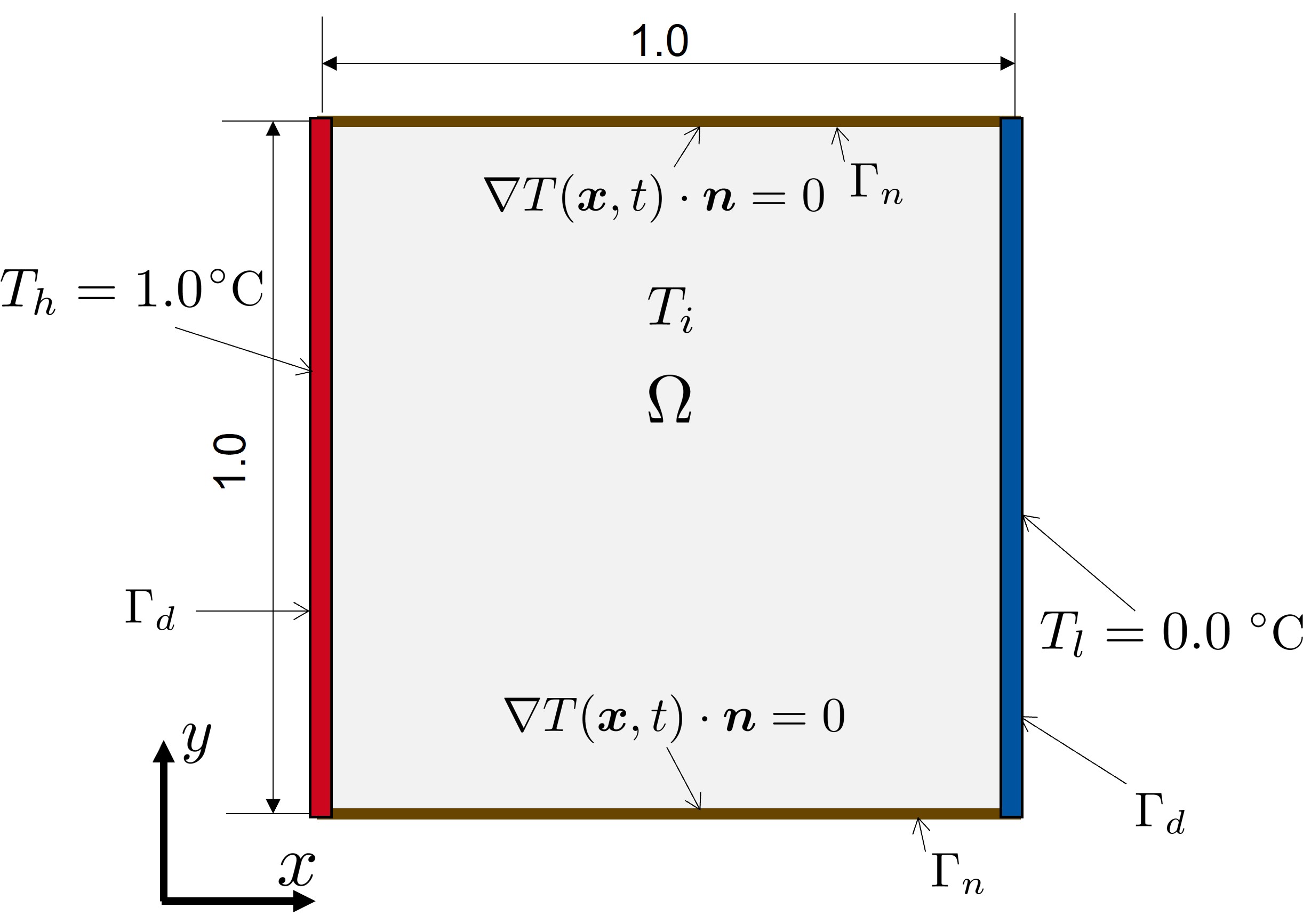}
    \caption{Dimensions of the problem domain and the boundary conditions.}
    \label{dimensions_bc}
  \end{minipage}
  \hfill
  \begin{minipage}{0.49\textwidth}
    \centering
    \includegraphics[width=70mm]{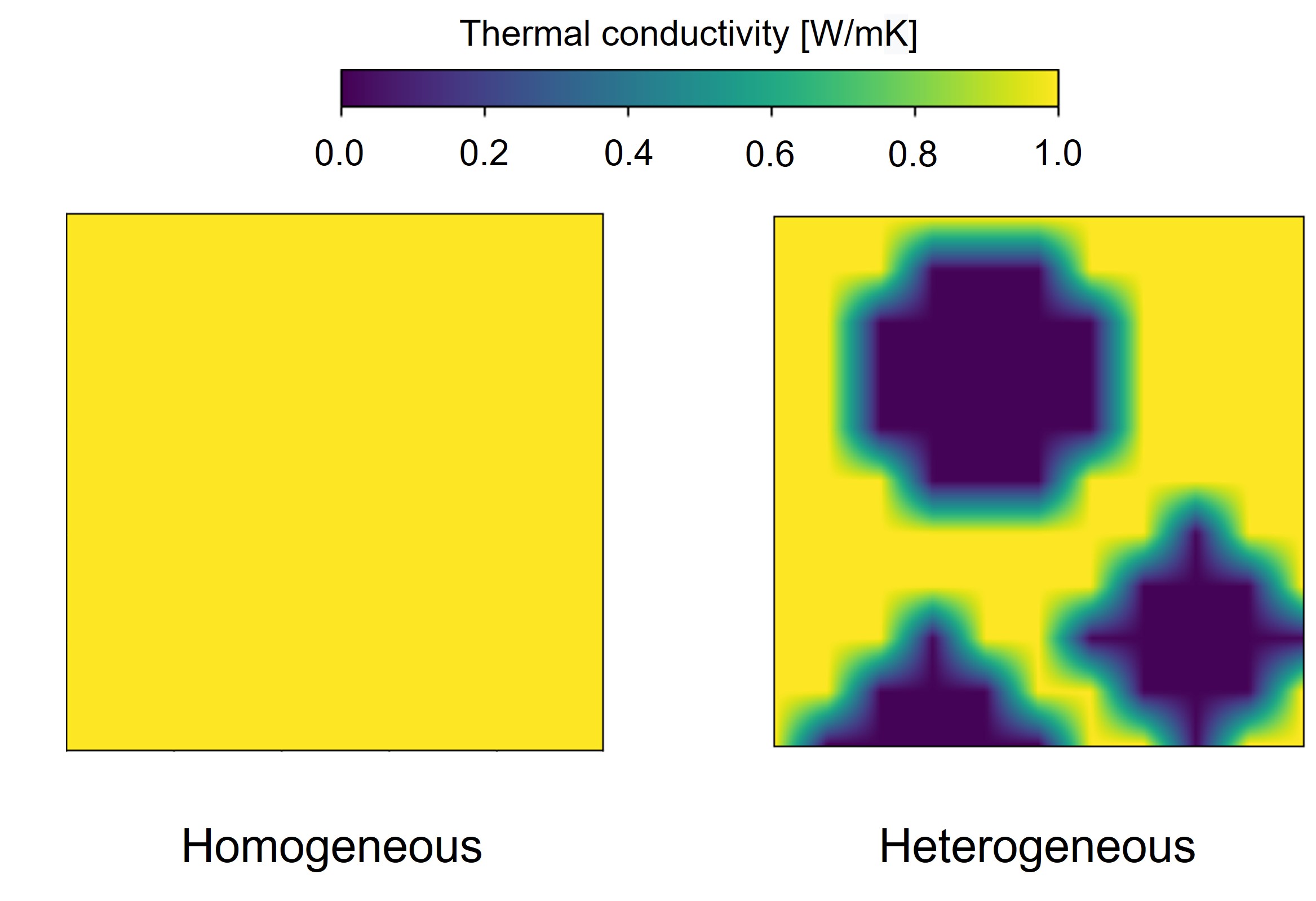}
    \caption{Two types of thermal conductivity distributions considered in this study.}
    \label{heat_cond_types}
  \end{minipage}
\end{figure}

\begin{figure}[t]
    \centering
    \includegraphics[width=150mm]{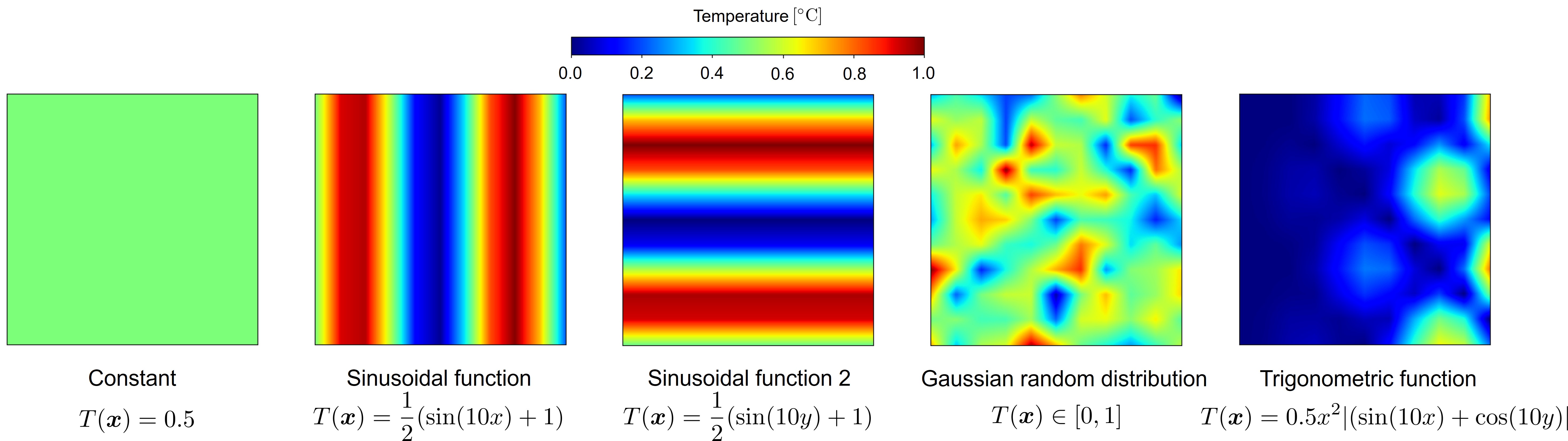}
    \caption{Five types of initial temperature fields for evaluating the performance of the trained networks. }
    \label{input_temp_four}
\end{figure}

The dimensions of the problem domain and the boundary conditions are depicted in Fig.~\ref{dimensions_bc}. One can imagine that a heat source supplies heat into the system from the left, and the right boundary is connected to a cold device that removes the heat from the system. In this problem setup, the heat source on the left boundary is prescribed as a Dirichlet boundary condition with a temperature of $1.0$ $^\circ$C. Similarly, the heat sink on the right boundary has a temperature of $0.0$ $^\circ$C. The Neumann boundary condition, which does not allow heat transfer, is applied to the top and bottom boundaries. 
This study considers two types of thermal conductivity distributions over the domain, homogeneous and heterogeneous, the distributions of which are shown in Fig.~\ref{heat_cond_types}. For instance, the microstructure of carbon fiber-reinforced plastics or architectural metamaterials can be used for heterogeneous thermal conductivity cases.
As initial temperature fields, five different distributions are conceived and considered in Fig.~\ref{input_temp_four} to test the performance of the network prediction for different temperature inputs. 
The initial temperature field, represented by an $11$ by $11$ grid of linearly discretized finite element points, is input into a neural network as described in Section 3.2. Physical fields, such as temperature fields and heterogeneity maps, are then upscaled to a $165$ by $165$ grid using bilinear shape functions. Further details on sample temperature fields for the training of the DL model are provided in Section 3.3. 
\subsection{Proposed finite element-based physics-informed operator learning framework}
\begin{figure}[t]
    \centering
    \includegraphics[width=140mm]{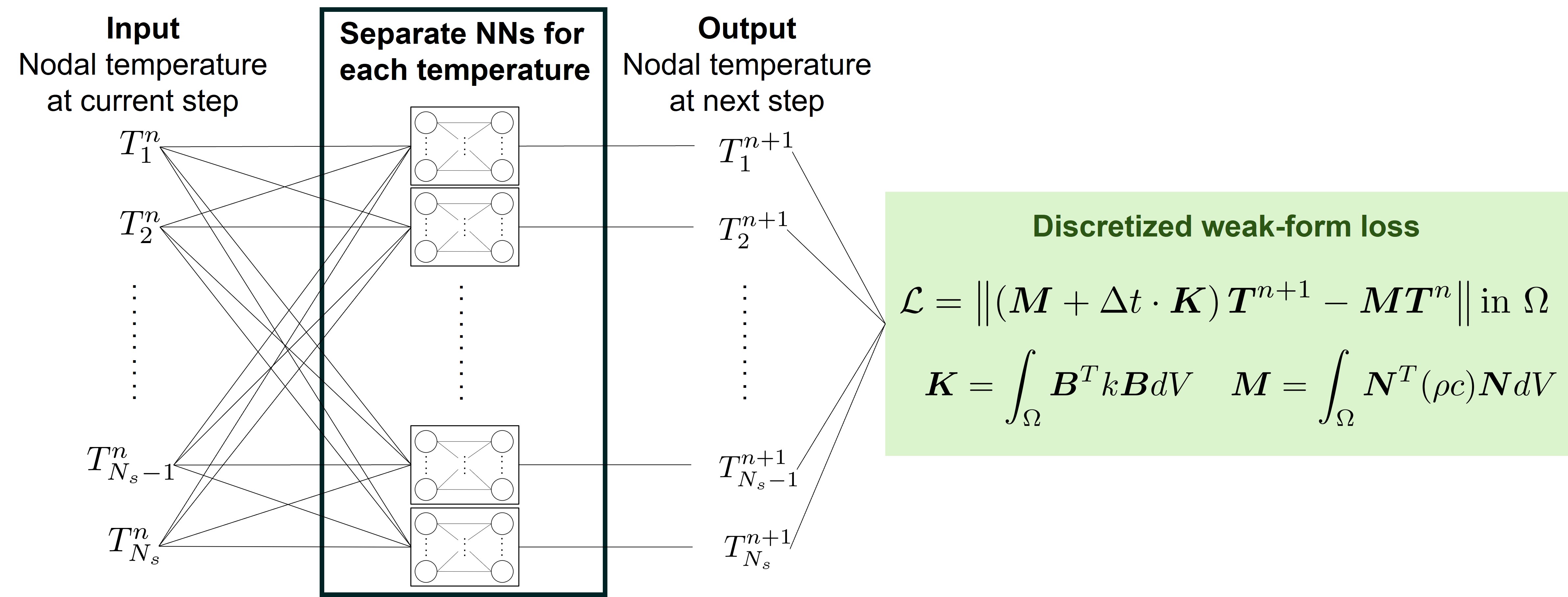}
    \caption{Network architecture and loss function used in the proposed framework.}
    \label{fig:network_architecture_loss}
\end{figure}

\begin{figure}[t]
    \centering
    \includegraphics[width=140mm]{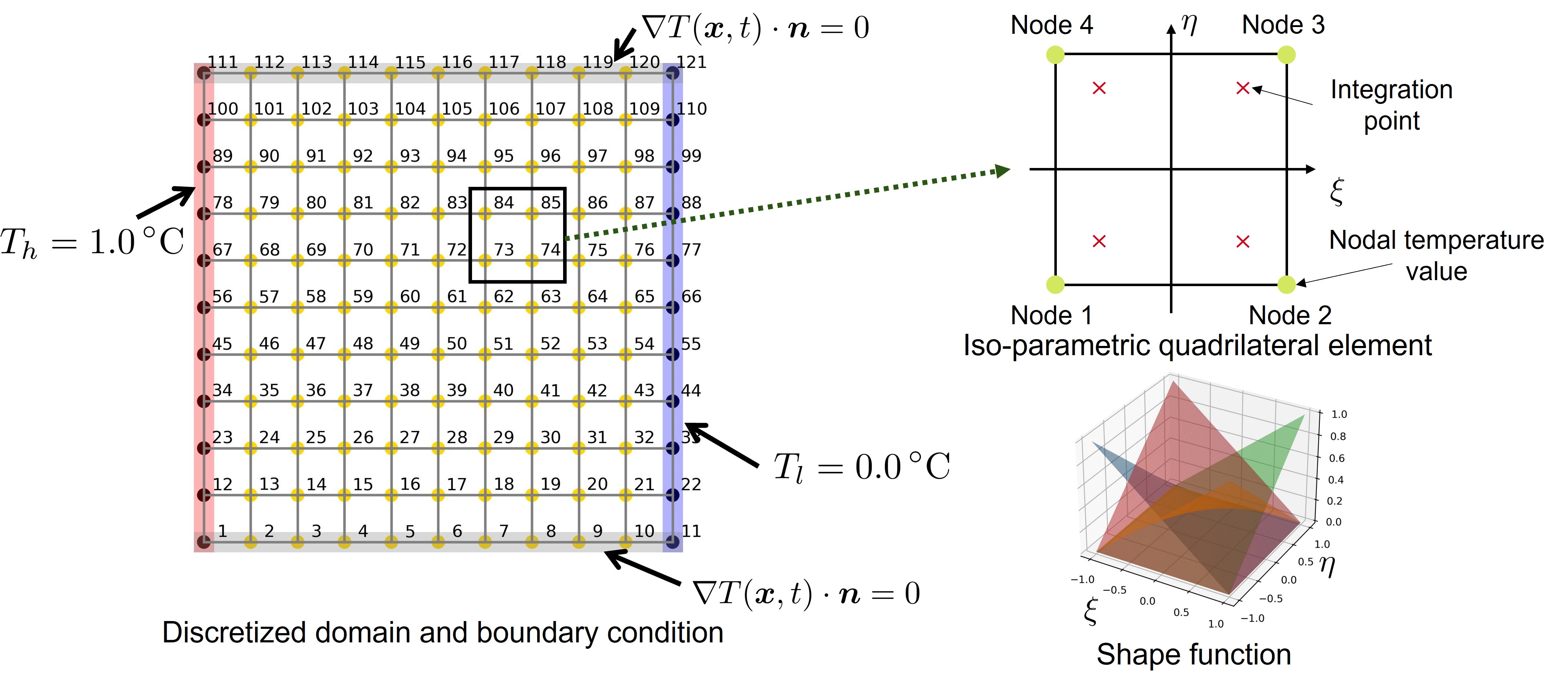}
    \caption{Discretized domain by finite elements. The networks evaluate the yellow nodes in the training and prediction stages. The black nodes are removed from the training target by applying hard boundary conditions.}
    \label{fig:grid}
\end{figure}

The core idea of the FOL framework is to predict physical fields at the next time step, utilizing their current time step state, which is equivalent to other time marching FE solvers. The network architecture is shown in Fig.~\ref{fig:network_architecture_loss}, implemented using the TensorFlow-based deep learning library SciANN \cite{Haghighat.2021}. The domain is first discretized through finite elements; see Fig.~\ref{fig:grid}. The nodes in the discretized domain are the representative points for evaluating temperature evolution by the networks. Analogous to the finite element method, the Gaussian integration is performed to integrate over the elements using the bi-linear shape function, shown in the right of Fig.~\ref{fig:grid}. Regarding the network architecture, it is worth noting that separate feedforward NNs are used to predict each node's temperature output. In \cite{rezaei2024integration}, the authors showed that separate networks with a small number of neurons per layer in each network outperformed a single fully connected network with a large number of neurons per layer. 
Nevertheless, it is also shown in the same work that using a simple fully connected network with a properly reduced architecture performs very well in finding the correct solutions. Therefore, the user needs to study this matter according to the problem at hand and the nature of the equations and outputs.
The comparison of the performance between the separated network architecture and the fully connected architecture is described in Section 4.5.
All the separated NNs are trained at the same time through a physically informed loss function based on the input and output temperature fields. 
Substituting $\alpha = 1 $, which means backward Euler approximation in time, into Eq.~\ref{eq:FE_heat} and taking the L2 norm of the residual yields the loss function in this framework, which reads, 
\begin{equation}
\label{eq:totalloss}
    \mathcal{L} =\left\|\left ( \bm M + {\Delta t}  \bm K \right) \bm T^{n+1}- \bm M \bm T^{n}  \right\| \ \text{in} \ \Omega ,
\end{equation}
where $\| \cdot \|$ denotes the L2 norm. More concretely, $\bm M$ and $\bm K$ are constructed as,
\begin{equation}
    \bm M = \mathrm{\bm{\mathcal A}}_{e=1}^{n_{e l}}\sum_{j = 1}^{n_{gauss}} \bm N^T_e(\bm{\xi}_j)~\rho c~\bm N_e(\bm{\xi}_j) \det{J(\bm{\xi}_j)}~\mu_j,  
\end{equation}
\begin{equation}
    \bm K = \mathrm{\bm{\mathcal A}}_{e=1}^{n_{e l}}\sum_{j = 1}^{n_{gauss}} \bm B^T_e(\bm{\xi}_j)~k(\bm{\xi}_j)~\bm B_e(\bm{\xi}_j) \det{J(\bm{\xi}_j)}~\mu_j. 
\end{equation}
Here, $\mathrm{\bm{\mathcal A}}_{e=1}^{n_{e l}}$ denotes the assembly of the element contributions from element $1$ to element $n_{el}$ (total number of elements), $n_{gauss}$ is the number of Gaussian points, $\bm{\xi}_j$ is the coordinate of $j$-th Gaussian point, and $\mu_j$ is the weight of Gaussian quadrature for $j$-th Gaussian point. In Eq.\,\ref{eq:totalloss}, $\bm{T}^n$ and $\bm{T}^{n+1}$ can be considered the input and output temperature fields of the network, respectively (the initial temperature field as well as the next step one). 

To enforce Dirichlet boundary conditions, this framework employs hard-constrained boundary conditions for the nodes, focusing solely on predicting unknown temperatures. This procedure is once again very similar to the classical finite element approach. In Fig.~\ref{fig:grid}, only the inner nodes, colored yellow, are evaluated and predicted through the networks. On the other hand, the black-colored nodes at the left and right boundaries are removed from the set of nodes used for training. However, it is worth mentioning that the influence of the Dirichlet boundary nodes is taken into account through the formulation of the physics-informed loss function in which the nodal field values of the Dirichlet boundary nodes are incorporated. 

In the training, mini-batch learning with multiple input samples is employed to optimize the networks. The loss in each mini-batch iteration is defined with the mean squared error as
\begin{equation}
    \mathcal{L} = \frac{1}{n_s} \sum_{i=1}^{n_s} \mathcal{L}_i^2,
\end{equation}
where $n_s$ is the number of samples per mini-batch.
The predictive performance by the trained networks is evaluated by the relative L2-norm error $E_{rr}$, which reads,
\begin{equation}
    E_{rr} = \frac{\| \bm T_{NN} - \bm T_{FE} \|}{\| \bm T_{FE} \|}.
\label{eq:E_rr}
\end{equation}
In Eq.~\ref{eq:E_rr}, $\bm T_{NN}$ is the temperature predicted by the NNs and $\bm T_{FE}$ is the corresponding FE solution, both of which are stored in a vector form. 
Finally, the networks are trained with this setup. A list of the hyperparameters of the networks, as well as the material parameters used in the study, are summarized in Table \ref{Table:hyperparameters}. For the units of the material parameters and the temperature, we consider Kelvin as a unit for temperature; here we consider the temperature ranges from 0 $^\circ$C to 1 $^\circ$C, and as for the thermal conductivity, W/mK is assumed.

\subsection{Generation of training samples}
\begin{figure}[H]
  \begin{minipage}{.50\linewidth}
    \centering
    \captionof{table}{Summary of hyperparameters and material parameters}
 \begin{tabular}{ccc} \\
    \hline
        Training parameter &   Value\\
    \hline
     Number of epochs & 1000,3000,5000 \\
     Optimizer& Adam, L-BFGS \\
     Grid structure & $11 \times 11$, $21 \times 21$ \\
     Structure of each NN & [10, 10]\\
     Activation function & Swish, Tanh, Sigmoid, ReLU\\
     Number of samples & 1000, 3000, 5000\\
     Learning rate & 0.001 \\
     Batch size & 60, 100 \\
     Time step size $\Delta t$ &0.01, 0.025, 0.05 [s]\\
    \hline 
    \hline 
     Density $\rho$ &10.0 [kg/m$^3$]\\
     Heat capacity $c$ &1.0 [J/kg$\cdot$K]\\
    \hline\\    
    \end{tabular}
    \label{Table:hyperparameters}
  \end{minipage}
    \begin{minipage}{.5\linewidth}
    \centering
    \includegraphics[width=50mm]{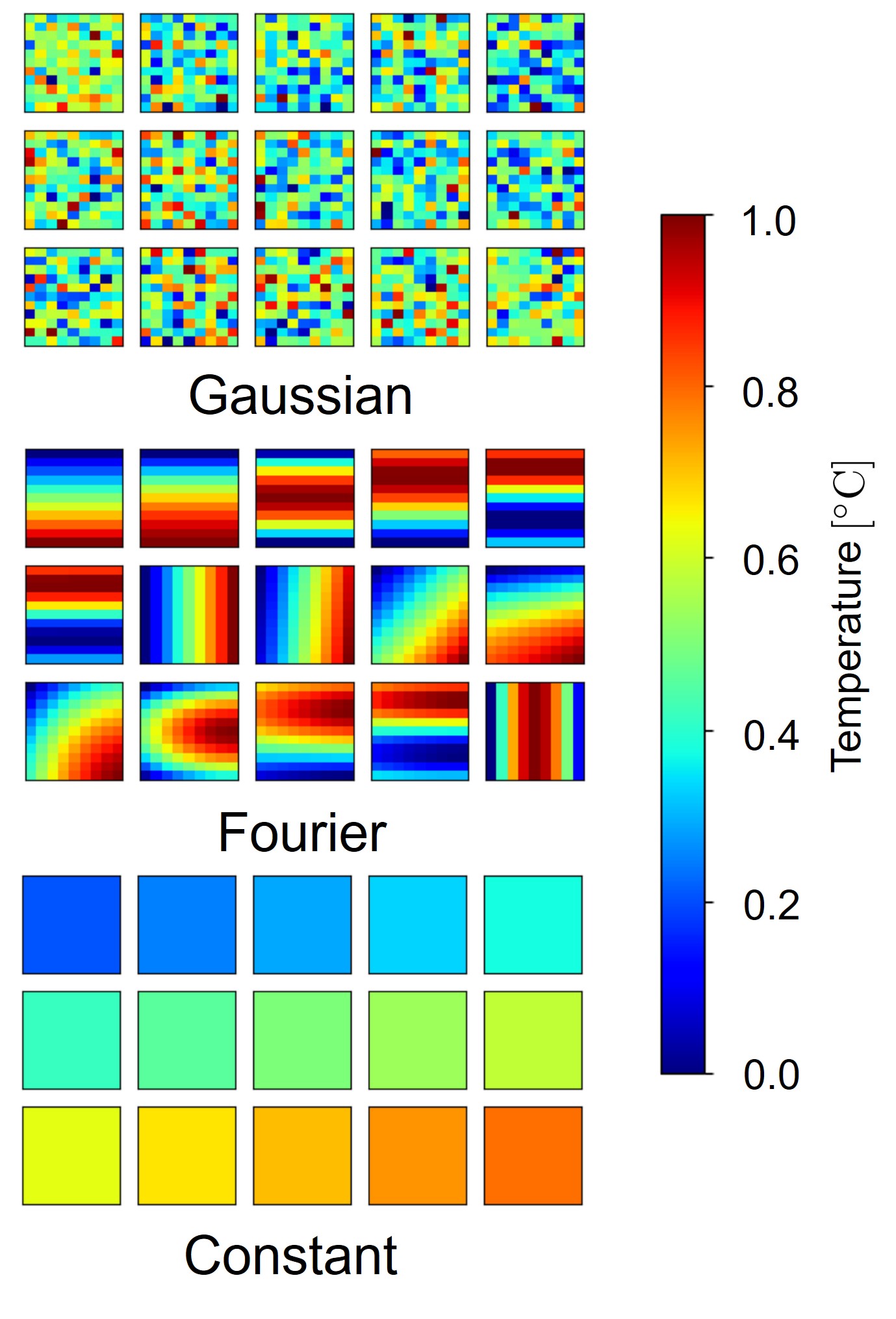}
    \captionof{figure}{Examples of input temperature samples generated by the Gaussian, Fourier, and constant-temperature generators. }
    \label{fig:gen_input_samples}
  \end{minipage}
\end{figure}

The input samples used for training are generated by combining three types of functions, i.e.,  Gaussian distribution, Fourier series, and constant field. Some of the generated input samples are shown in Fig.~\ref{fig:gen_input_samples}. The first is based on the Fourier series. The temperature field generated by the Fourier series with randomly generated amplitudes and frequencies has a smooth distribution without a steep gradient. 
The function is given as,
\begin{equation}
\label{eq:foruierGen}
\begin{aligned}
    T(\bm x) &= \sum_i^{n_{sum}} [c_i + A_i \sin{(C_i\cdot x)} \cos{(D_i\cdot y)} +B_i \cos{(C_i\cdot x)} \sin{(D_i\cdot y)} \\ & + A_i\sin{(C_i\cdot x)} \sin{(D_i\cdot y)} +B_i \cos{(C_i\cdot x)} \cos{D_i\cdot y)}].
\end{aligned}
\end{equation}
In Eq.\,\ref{eq:foruierGen}, $\bm x = (x, y)^T$, $c_i$ is the real-valued constant, $A_i$ represents the amplitude in the $x$-direction, $B_i$ the amplitude in the $y$-direction, $C_i$ represents the frequency in the $x$-direction, and $D_i$ the frequency in the $y$-direction. 
These are parameters for generating different patterns of temperature distribution. The ranges of values from which the parameters $c_i, A_i, B_i, C_i, \ \text{and} \ D_i$ are randomly chosen, are determined (see Table \ref{Table:Fourier_param}):
\begin{equation}
\begin{aligned}
    c_i & \in c_r = \left\{ r  \ | \ a_c\le  r \le  b_c\right\} \\
    A_i & \in A_r = \left\{ r  \ | \ a_A\le  r \le  b_A\right\} \\
    B_i & \in B_r = \left\{ r  \ | \ a_B\le  r \le  b_B\right\} \\
    C_i & \in C_r = \left\{ r  \ | \ a_C\le  r \le  b_C\right\}\\
    D_i & \in D_r = \left\{ r  \ | \ a_D\le  r \le  b_D\right\}. \\
\end{aligned}
\end{equation}

For this training $n_{sum}$ was set to $50$. 
The parameters are generated $n_{sum}$ times and then the input temperature samples are created by summing the Fourier series $n_{sum}$ times with the prepared parameter set. At the end, a normalization is performed to restrict the range between $0$ and $1$. 
In total, $1200$ input samples were prepared using this procedure in this study.
The second temperature generator comes from the Gaussian random process.  
For a generation of input samples, the output is normalized between 0 and 1 after initial random patterns are generated. This process is done iteratively for each node and the number of input samples by the Gaussian generator. With this generator, 1500 input samples were prepared for training.
In addition to the input samples generated by the two temperature generators, one can also consider input samples with varying constant temperatures to increase the variety of training data. This generator created 300 input samples for the training. 
In total, 3000 input samples were eventually generated from the three types of functions to cover a wide range of temperatures considered in the training process.  

\begin{table}[H]
\centering
\caption{Parameters used for generating temperature samples with the Fourier series.}
 \begin{tabular}{ccc} \\
    \hline
        Parameter &   Value\\
    \hline
     $(a_c,b_c)$ & $\left\{(0.0, 0.5), (0.5, 1.0), (1.0, 1.5)\right\}$\\
     $(a_A,b_A)$& $\left\{(0.01, 0.1), (0.1, 0.5), (0.5, 1.0), (1.0, 1.5), (1.5, 2.0)\right\}$\\
     $(a_B,b_B)$ & $\left\{(0.01, 0.1), (0.1, 0.5), (0.5, 1.0), (1.0, 1.5), (1.5, 2.0)\right\}$ \\
     $(a_C,b_C)$ & $\{(0.0, 0.0), (0.001, 0.01),(0.01, 0.1),(0.1, 1.0),(1.1, 2.0), (2.1, 4.0), (4.1, 6.0)\}$\\
     $(a_D,b_D) $ & $\{(0.0, 0.0), (0.001, 0.01),(0.01, 0.1),(0.1, 1.0),(1.1, 2.0), (2.1, 4.0), (4.1, 6.0)\}$\\
    \hline\\    
    \end{tabular}
    \label{Table:Fourier_param}
\end{table} 
\section{Results}
\begin{figure}[t]
    \centering
    \includegraphics[width=120mm]{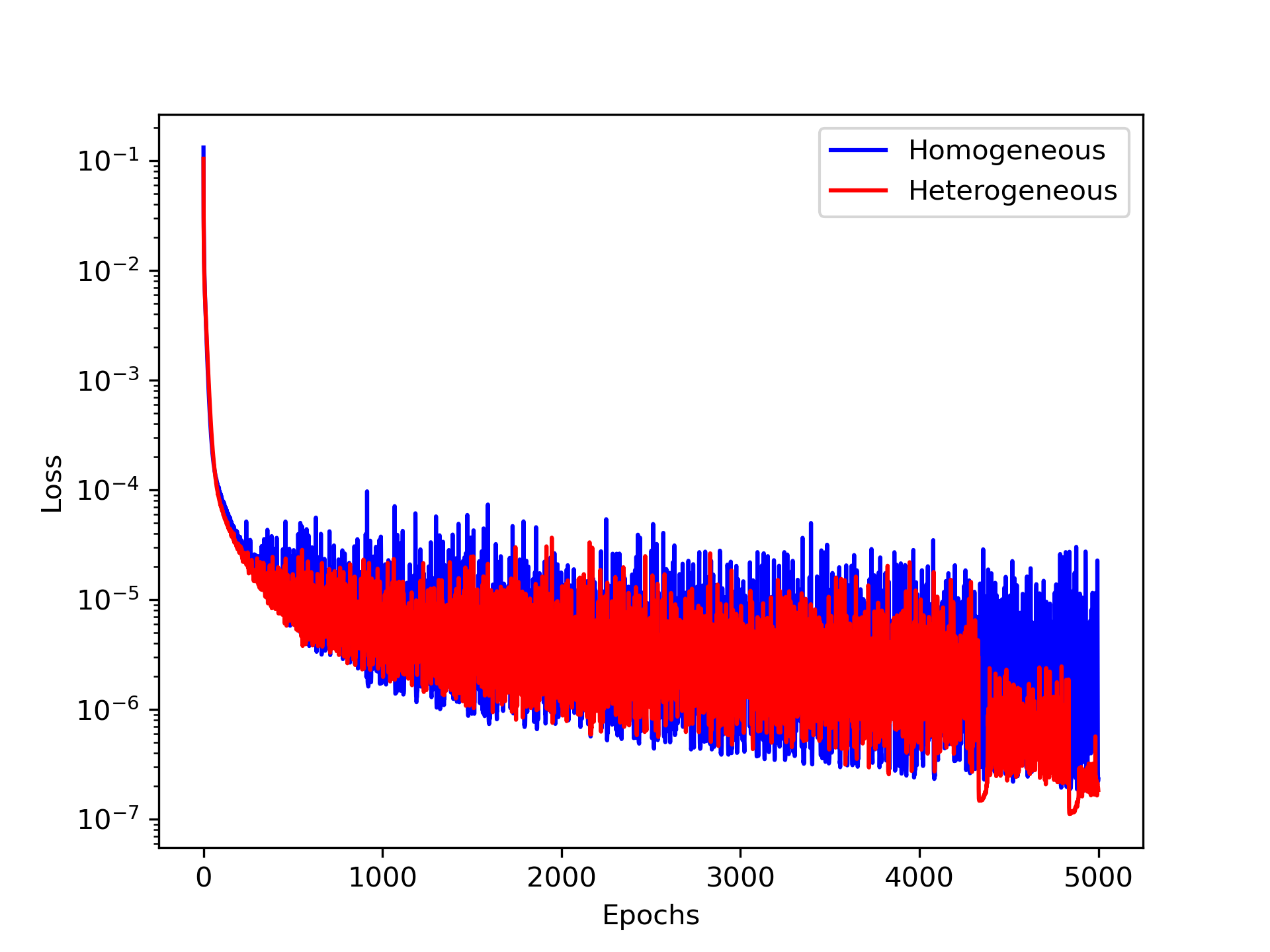}
    \caption{Loss history for the homogeneous and heterogeneous thermal conductivities.}
    \label{fig:loss_history}
\end{figure}

\begin{figure}[H]
    \centering
    \includegraphics[width=160mm]{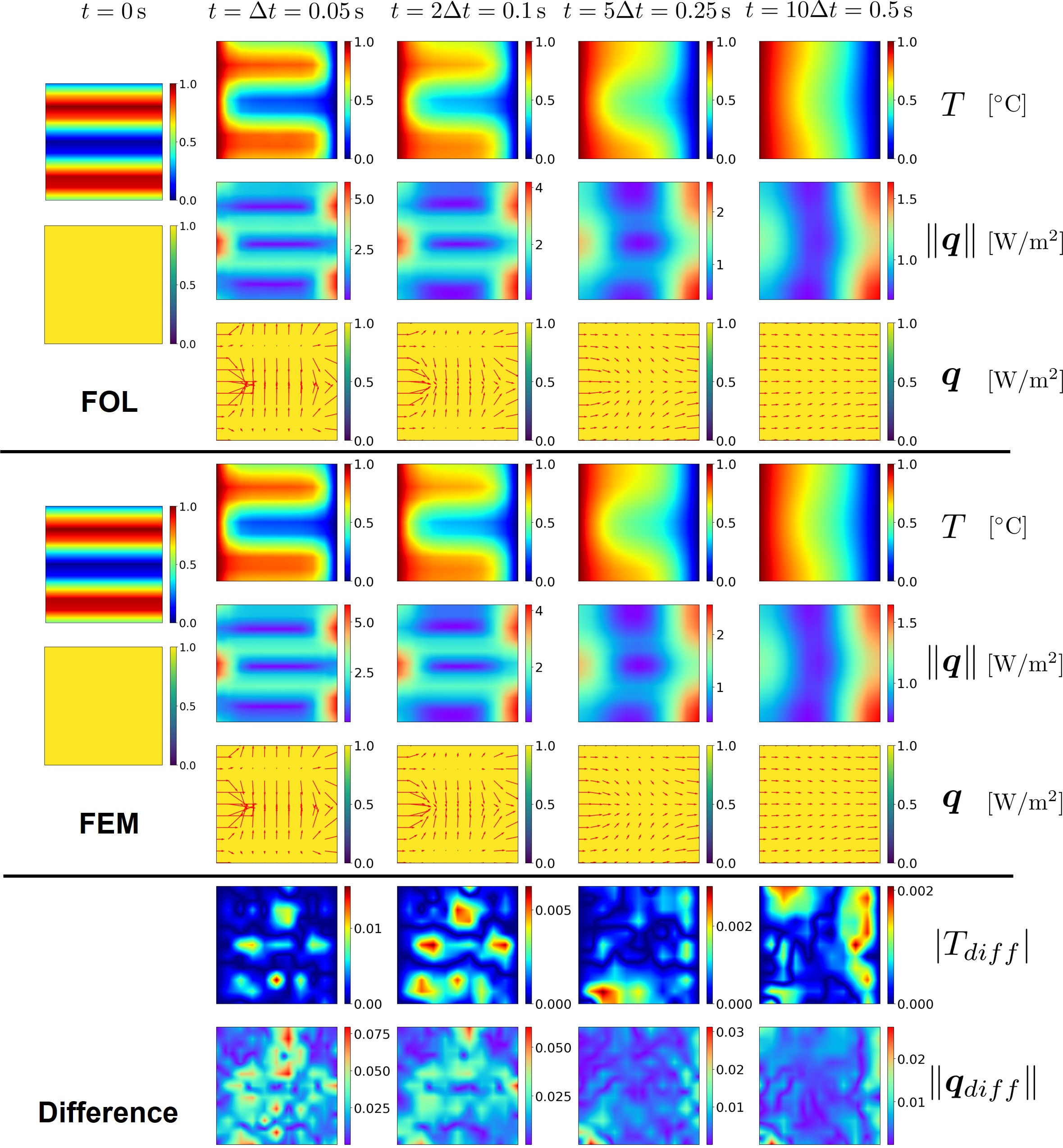}
    \caption{Temperature predictions and obtained heat flux by the networks (top), reference solutions by FEM (middle), and the difference between the predictions and reference solutions (bottom) in the case with the homogeneous thermal conductivity and temperature field with $T(\bm x) = \frac{1}{2}\left(\sin(10 y) + 1 \right)$ as an initial temperature field.}
    \label{homoge_sin}
\end{figure}

\begin{figure}[H]
    \centering
    \includegraphics[width=160mm]{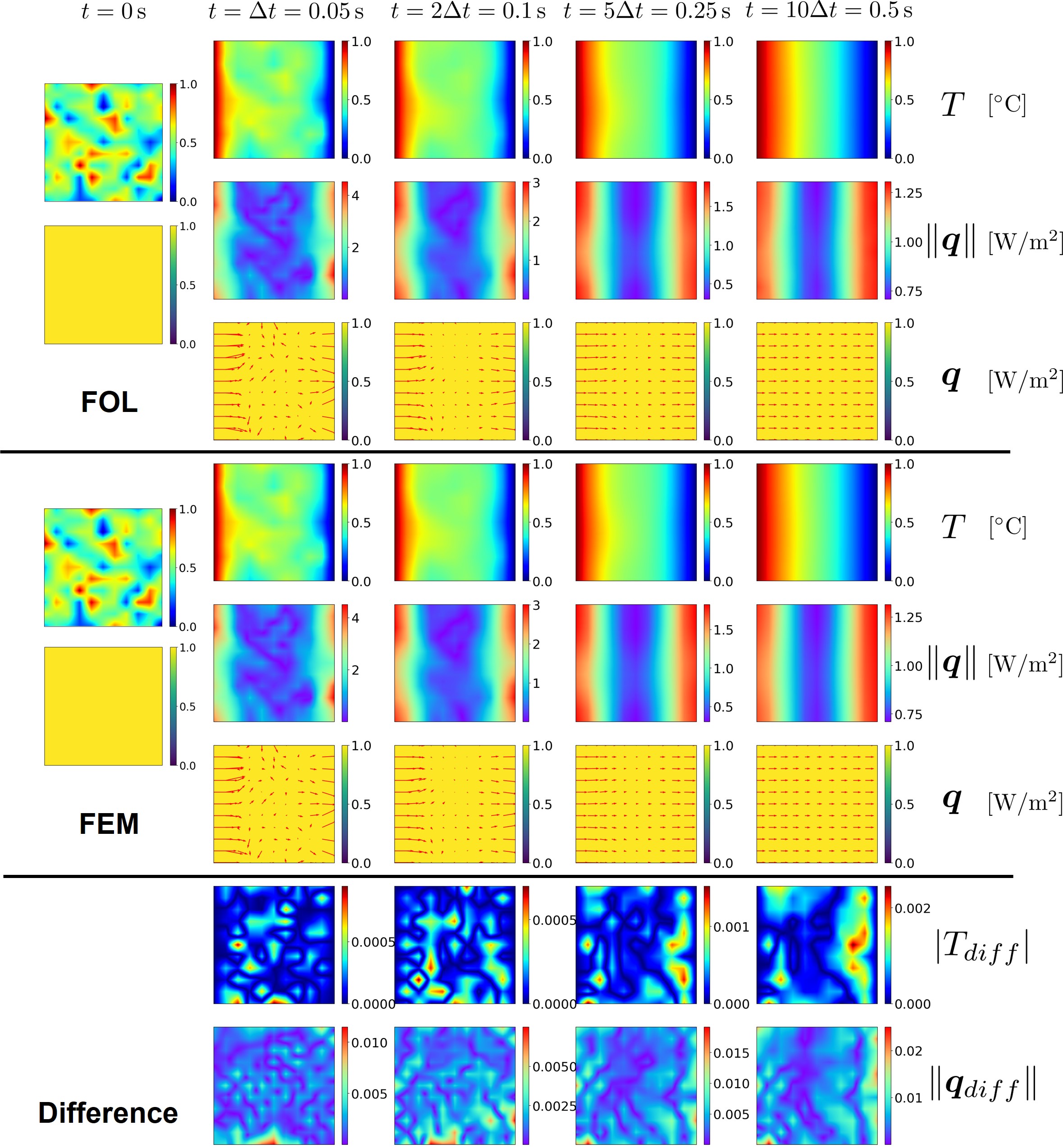}
    \caption{Temperature predictions and obtained heat flux by the networks (top), reference solutions by FEM (middle), and the difference between the predictions and reference solutions (bottom) in the case with the homogeneous thermal conductivity and temperature field with the Gaussian distribution-based temperature field as an initial temperature field.}
    \label{homoge_gauss}
\end{figure}

\begin{figure}[H]
    \centering
    \includegraphics[width=160mm]{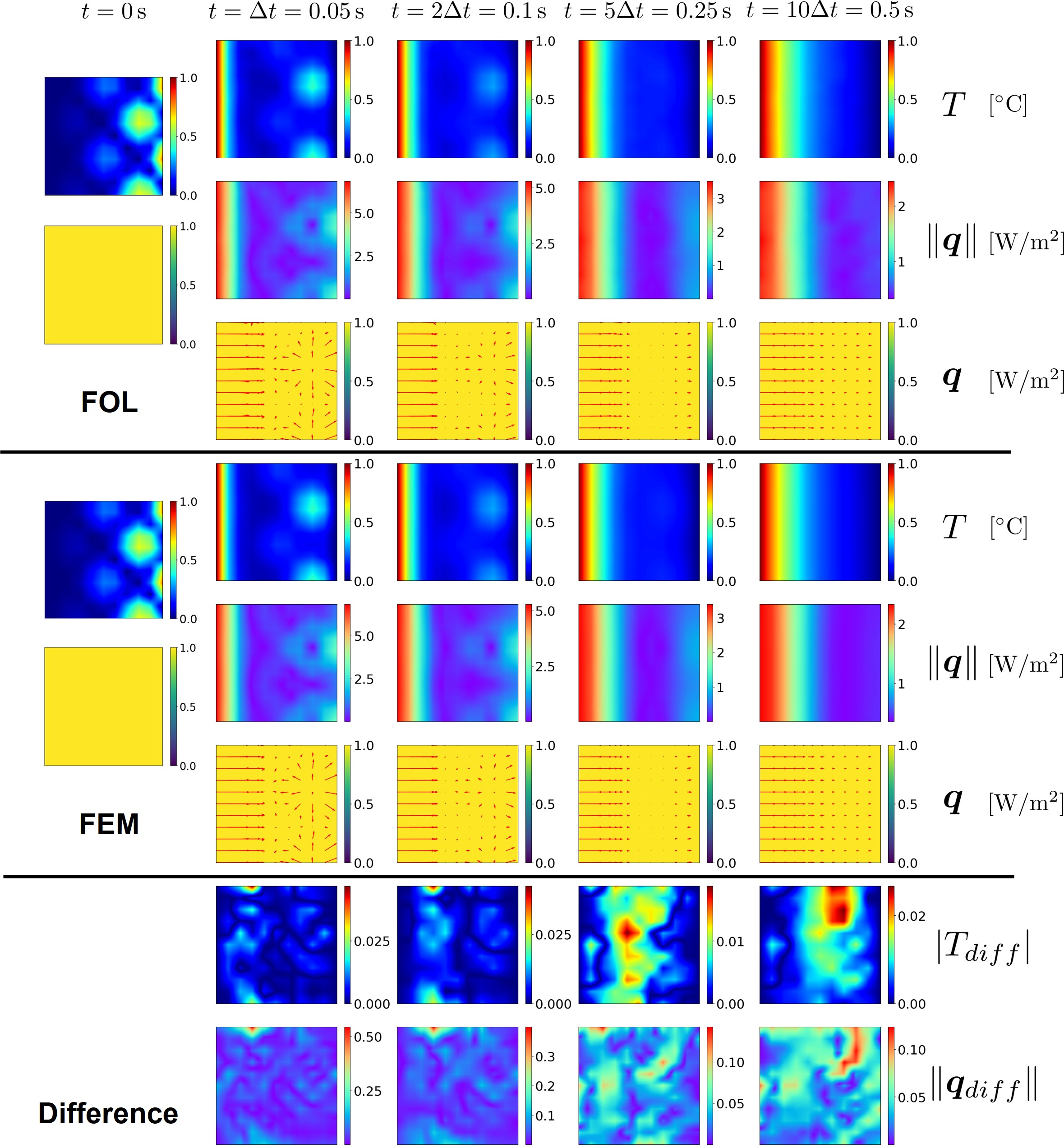}
    \caption{Temperature predictions and obtained heat flux by the networks (top), reference solutions by FEM (middle), and the difference between the predictions and reference solutions (bottom) in the case with the homogeneous thermal conductivity and temperature field with $T(\bm x) = 0.5x^2 |(\sin(10x)+\cos(10y)|$ as an initial temperature field.}
    \label{homoge_trig}
\end{figure}

\begin{figure}[H]
    \centering
    \includegraphics[width=160mm]{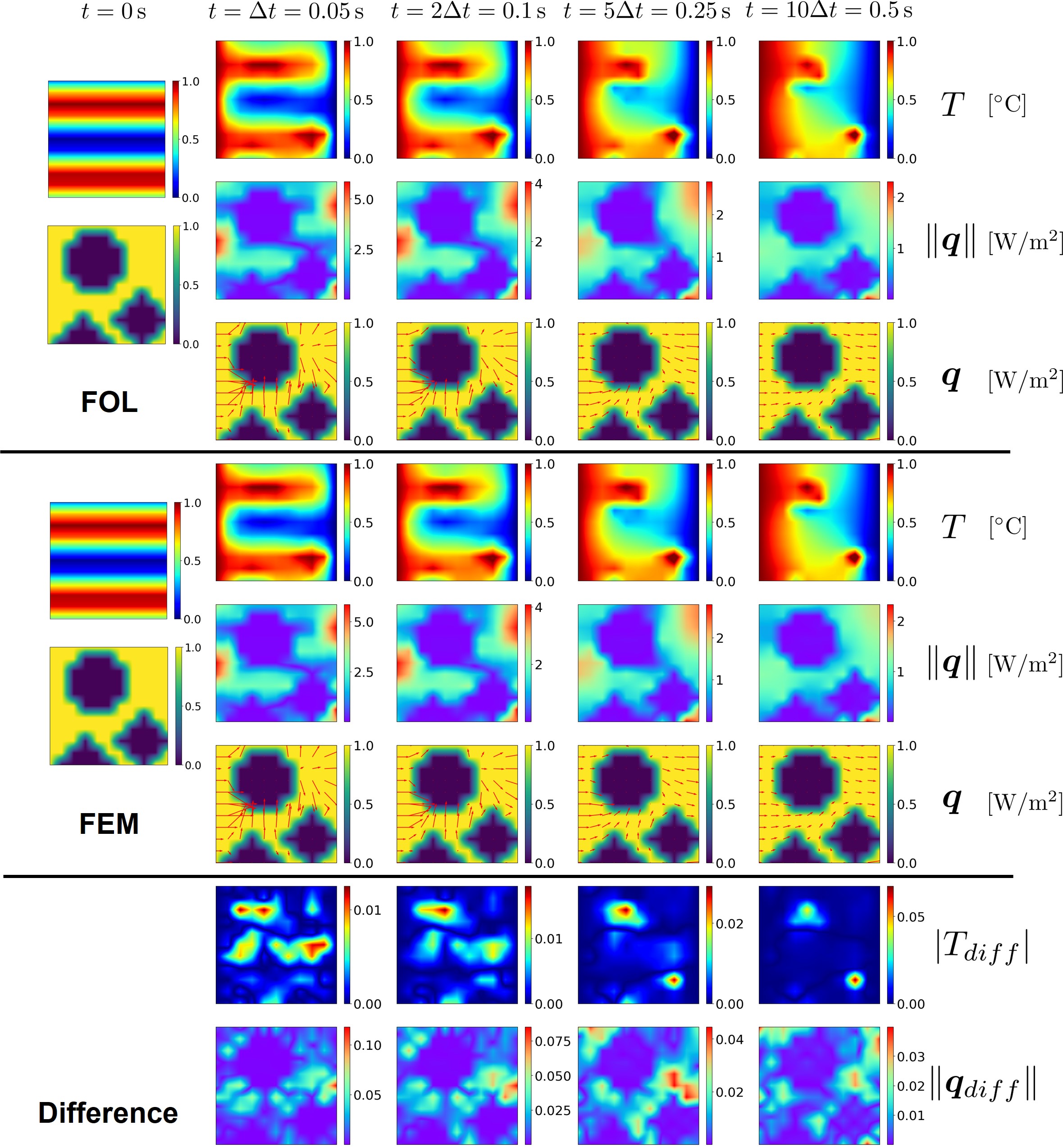}
    \caption{Temperature predictions and obtained heat flux by the networks (top), reference solutions by FEM (middle), and the difference between the predictions and reference solutions (bottom) in the case with the heterogeneous thermal conductivity and sinusoidal temperature field with $T(\bm x) = \frac{1}{2}\left(\sin(10 y) + 1 \right)$ as an initial temperature field. }
    \label{hetero_sin}
\end{figure}

\begin{figure}[H]
    \centering
    \includegraphics[width=160mm]{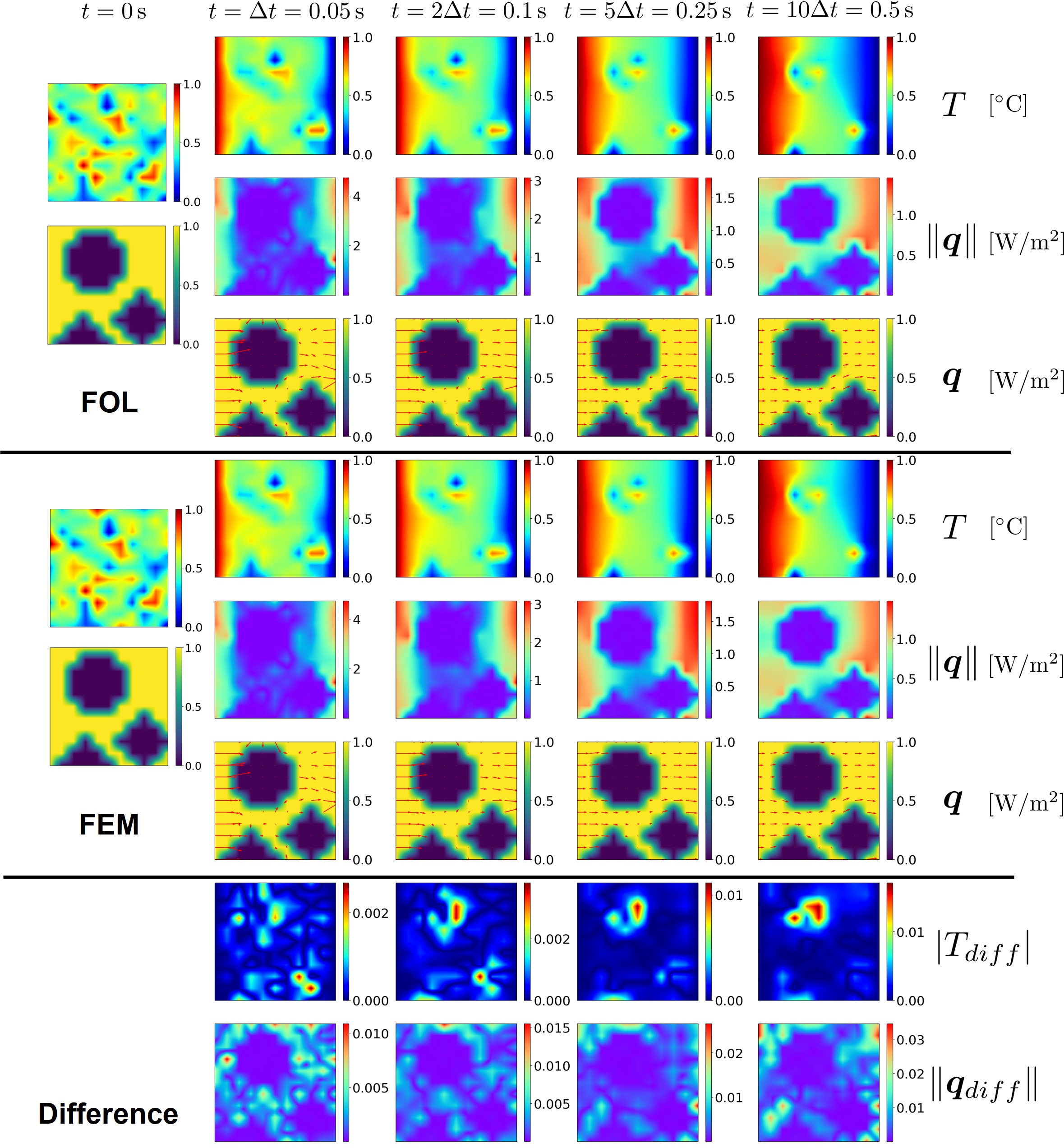}
    \caption{Temperature predictions and obtained heat flux by the networks (top), reference solutions by FEM (middle), and the difference between the predictions and reference solutions (bottom) in the case with the heterogeneous thermal conductivity and temperature field with the Gaussian distribution-based temperature field as an initial temperature field.}
    \label{hetero_gauss}
\end{figure}

\begin{figure}[H]
    \centering
    \includegraphics[width=160mm]{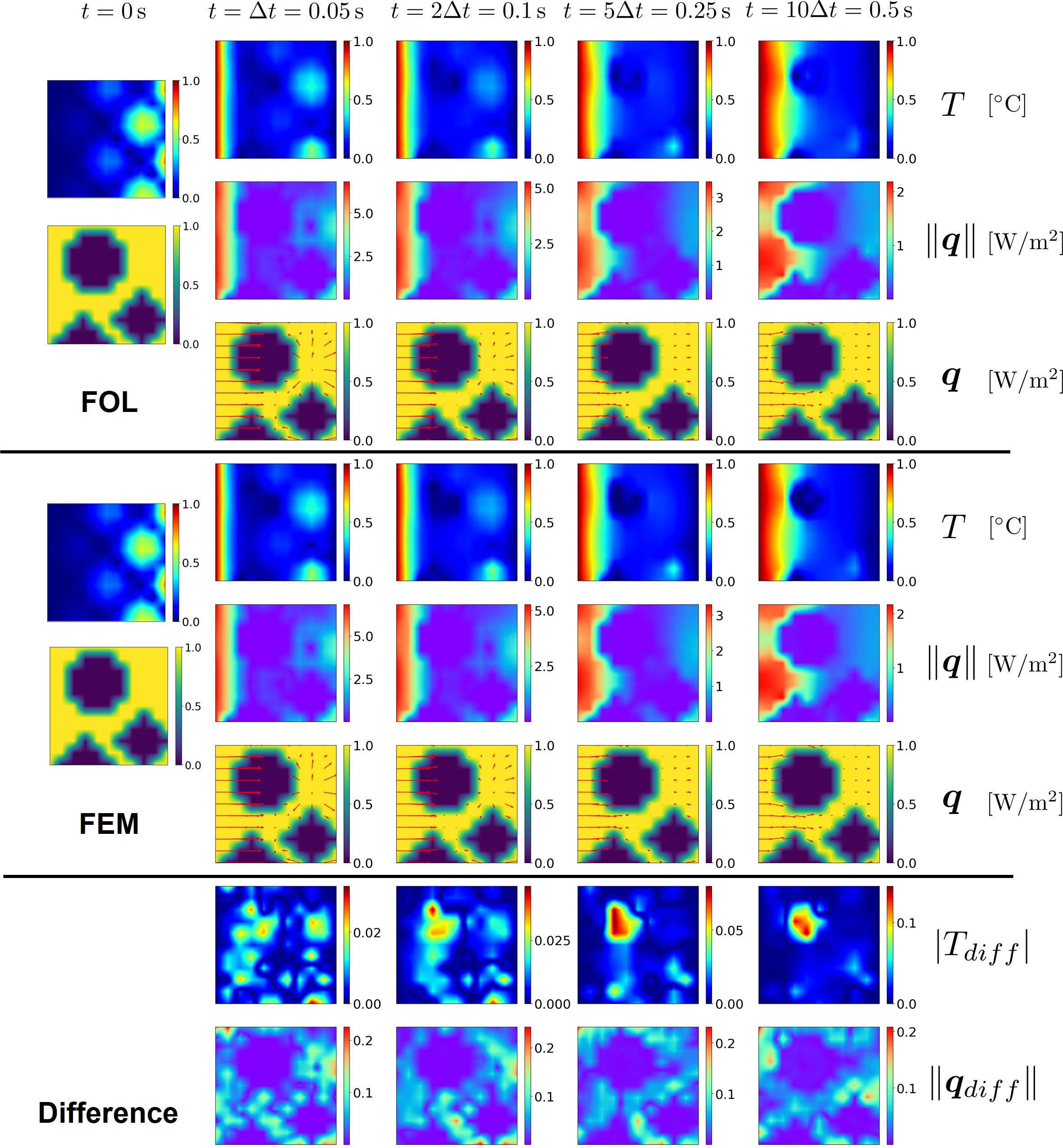}
    \caption{Temperature predictions and obtained heat flux by the networks(top), reference solutions by FEM (middle), and the difference between the predictions and reference solutions (bottom) in the case with the heterogeneous thermal conductivity and temperature field with $T(\bm x) = 0.5x^2 |(\sin(10x)+\cos(10y)|$ as an initial temperature field.  }
    \label{hetero_trig}
\end{figure}

\begin{figure}[H]
    \centering
    \includegraphics[width=150mm]{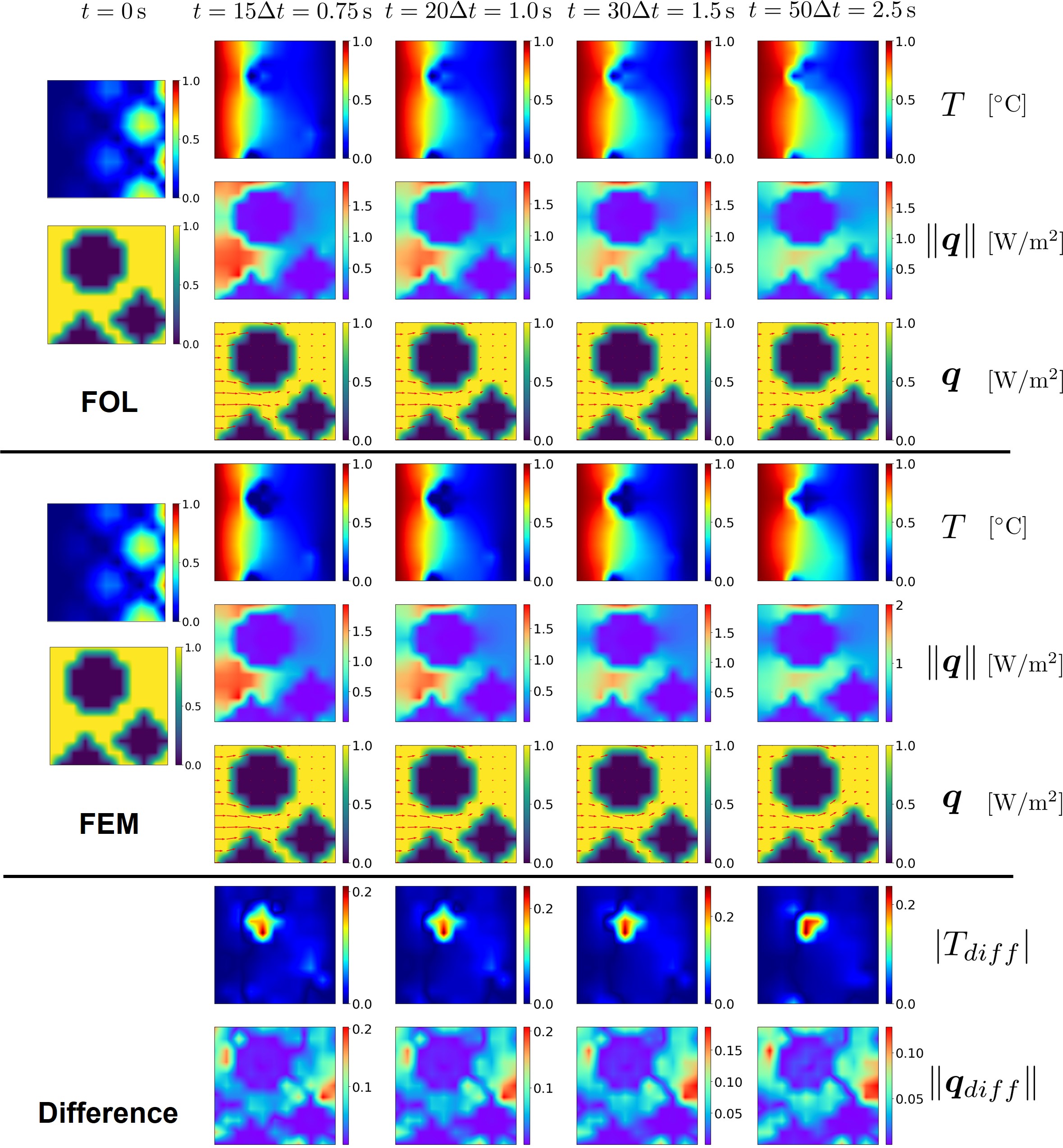}
    \caption{Temperature predictions and obtained heat flux up to $t = 50 \Delta t = 2.5$ by the networks (top), reference solutions by FEM (middle), and the difference between the predictions and reference solutions (bottom) in the case with the heterogeneous thermal conductivity and temperature field with $T(\bm x) = 0.5x^2 |(\sin(10x)+\cos(10y)|$ as an initial temperature field.}
    \label{hetero_trig_more}
\end{figure}

Using the framework introduced in the previous section, we have attempted to predict the solution of the heat equation for any given initial temperature field. As a demonstration, we show here only the results of the predictions for three of the initial temperature fields, one with $T(\bm x) = \frac{1}{2}\left(\sin(10 y) + 1 \right)$, one from the Gaussian distribution, and the last one with $T(\bm x) = 0.5x^2 |(\sin(10x)+\cos(10y)|$, considered in this work. We can confirm that the chosen initial temperatures are not exactly represented in the initial training samples. Furthermore, as will be shown later, the temporal temperature evolution results in complex patterns that significantly differ from the training samples. It is important to note that since the training is entirely unsupervised, the network is not provided with solutions for these initial temperature samples. Hence, the subsequent results serve as rigorous tests to evaluate the network's performance. 
In the main study, we trained the NNs for $5000$ epochs to ensure that the loss reaches a plateau due to convergence after hundreds of epochs. 
This is confirmed in Fig.~\ref{fig:loss_history}, although it continues to decrease slightly over the epoch. We do not employ fixed criteria on the residual value for stopping the training, as this is currently not the main scope of this work.
In post-processing, we calculated the heat flux based on the obtained temperature fields with respect to the discretized system to further investigate the physical behavior of the predicted heat conduction. 
When the homogeneous thermal conductivity is applied, Fig.~\ref{homoge_sin} exhibits that the FOL predictions agree with the reference FE solution with a maximum error of about $0.003$ in the nodal absolute temperature error and $0.03$ in the nodal absolute heat flux magnitude error at $t = 10 \Delta t = 0.5$ [s] with $T(\bm x) = \frac{1}{2}\left(\sin(10 y) + 1 \right)$ as the initial temperature field, where in Fig.~\ref{homoge_sin} $T_{diff}$ and $\bm q_{diff}$ denote the difference in temperature and heat flux magnitude between the prediction and reference solution at each node, respectively. The heat flux transition over time shows that the transient behavior is accurately predicted by the proposed FOL framework upon training. 
The absolute error distribution shown at the bottom of Fig.~\ref{homoge_sin} looks rather random. The error can be further reduced by, for example, enhancing the variety of training samples, which will lead to better coverage of possible temperature fields in the network input.
Looking at the results in Fig.~\ref{homoge_gauss}, we confirmed that even for a random temperature pattern generated from the Gaussian distribution, one can obtain reasonable predictions with a maximum of $0.003$ in nodal absolute temperature error and $0.03$ in nodal absolute heat flux magnitude error at $t = 10 \Delta t = 0.5$ [s].
Here, we define "reasonable prediction" as the relative L2 error being small against the finite element solution, which means the predictions by FOL capture the important features of the temperature evolutions. More concretely, for $N=11$, if the relative L2 error in temperature prediction is less than 0.1, the FOL prediction is at reasonable accuracy for example.
In Fig.~\ref{homoge_trig}, the temperature error increased by a factor of $10$ to approximately $0.03$ at maximum in the nodal absolute temperature error, while the heat flux magnitude error increased by a factor of $5$ to about $0.15$. In Fig. 15 in the nodal absolute heat flux magnitude error at $t = 10 \Delta t = 0.5$  [s] with $T(\bm x) = 0.5x^2 |(\sin(10x)+\cos(10y)|$ as the initial temperature field.  Nevertheless, the prediction correctly captured the main feature of the temperature evolution seen in the reference solution by FEM in the qualitative comparison of the results from FOL and FEM. 
When it comes to the case with the heterogeneous thermal conductivity, one can also see in Figs.~\ref{hetero_sin}, \ref{hetero_gauss} and \ref{hetero_trig} the agreement of the FOL prediction with the corresponding FEM solutions. The main difference to the homogeneous case is that the error accumulation in the temperature field is dominant around the inserted low conductivity regions (i.e., phase boundaries) due to steep changes in the solution. To enhance the solution's quality, one can train the neural network with additional input neurons and utilize advanced optimizers, such as L-BFGS, with hyperparameter tuning to prevent getting trapped in local minima. 
It is important to note that the prediction of dynamic temperature evolution in the other part of the domain is reasonable.
One can also confirm in Figs. \ref{hetero_sin}, \ref{hetero_gauss}, and \ref{hetero_trig} that the red arrows representing the heat flux bypass the low conductivity region area in both FOL and FEM. 
Furthermore, we also predicted for more time steps up to $t = 50\Delta t = 2.5$  [s] to see if a long-term prediction is possible; see Fig. \ref{hetero_trig_more}. The prediction was still in good agreement with the FE solution even at $t = 50\Delta t$, although error concentration is seen in the low conductivity region. Overall, the results of the demonstration on the transient heat conduction problem show that the proposed FOL framework for spatiotemporal PDEs is capable of predicting the solution and even its spatial gradients under a given time step size and boundary conditions.

\begin{figure}[H]
    \centering
    \includegraphics[width=160mm]{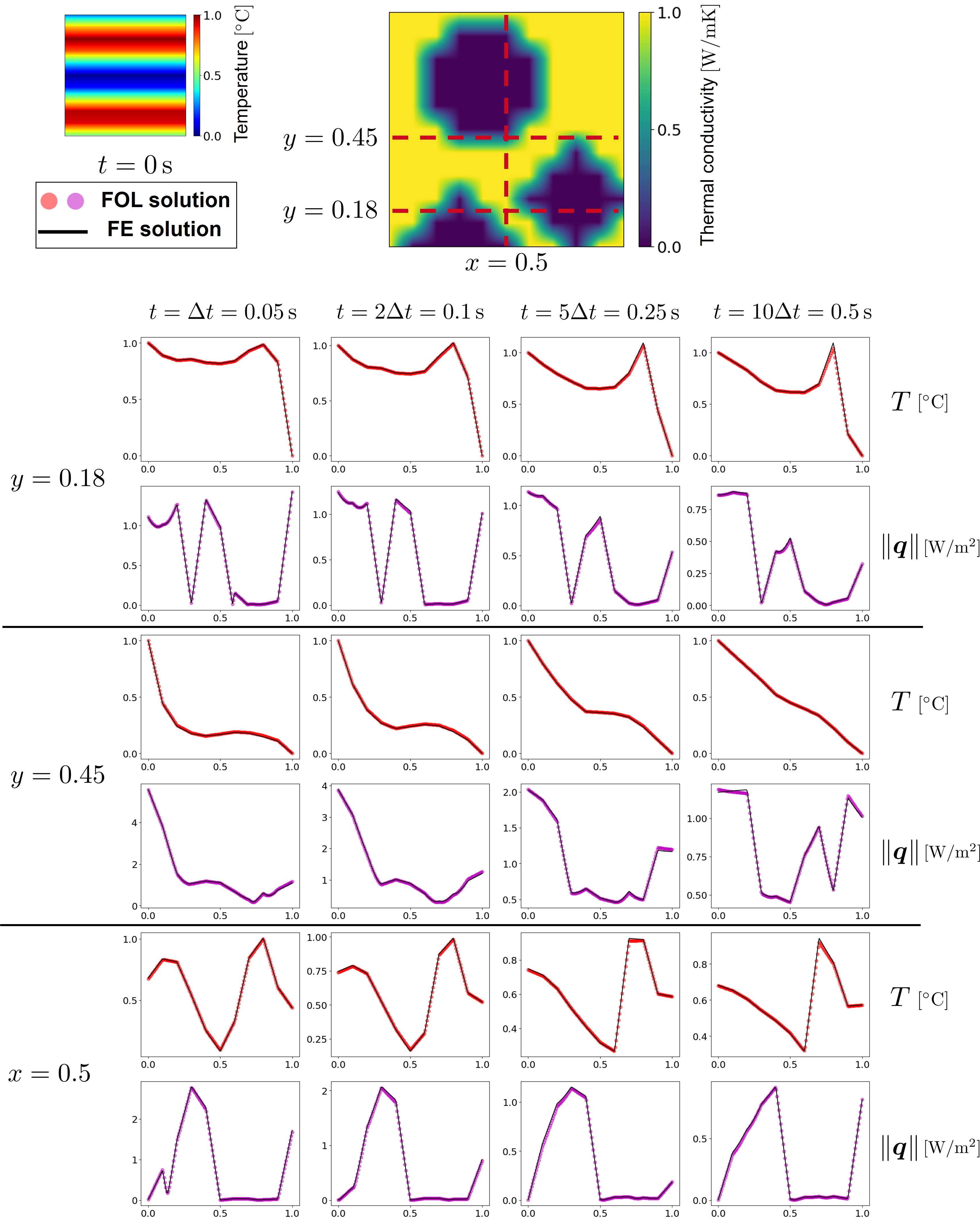}
    \caption{Cross-sections of temperatures and heat flux magnitudes at $y = 0.18$, $y = 0.45$, and $x = 0.5$ obtained from the networks and reference solutions by FEM in the case with the heterogeneous thermal conductivity and temperature field with $T(\bm x) = \frac{1}{2}\left(\sin(10 y) + 1 \right)$ as an initial temperature field. }
    \label{hetero_sin_cross}
\end{figure}

We also looked into cross-sectional changes in temperature and heat flux magnitude for the heterogeneous case with an initial temperature field of $T(\bm x) = \frac{1}{2}\left(\sin(10 y) + 1 \right)$ at $t = 10 \Delta t = 0.5$. As shown in Fig. \ref{hetero_sin_cross}, the predictions represented by the dots agree with the corresponding FE solutions, although one can find discrepancies with the FE solution at $t = 10 \Delta t$. This may also be due to the error accumulation problem. We conclude that after sufficient training with a variety of training samples, the proposed framework can predict the dynamic behavior of transient heat conduction with acceptable accuracy. 

\subsection{Influence of training samples}
This subsection examines the effect of the training samples on the prediction accuracy over time. The analysis includes not only the combination of Gaussian, Fourier, and constant fields but also the combinations of two out of the three temperature generators. To ensure that the constant fields account for only $10~\%$ of the total training samples, we determined the proportion of samples from each generator. For the dataset of the combination of Fourier and constant fields, we prepared $2700$ Fourier-type samples and $300$ constant-type samples. Similarly, for the dataset of the combination of Gaussian and constant fields, we prepared $2700$ Gaussian-type samples and $300$ constant-type samples. To observe the impact of sample size on prediction, a new case was added with $1000$ training samples, each with the same percentage as the original one. For this study, the NNs were trained for $1000$ epochs with $\Delta t = 0.05$  [s]. We also kept this condition for the other studies in the following subsections as well. The results are shown in Figs.~\ref{err_samples_homoge} and \ref{err_samples_hetero} for the homogeneous and heterogeneous thermal conductivities, respectively. Notably, the combination of Fourier and constant resulted in errors about twice as large as in the other three cases, including those obtained from $1000$ samples. This suggests that the accuracy of the prediction increases as more of the possible temperature field patterns are covered by training samples. The comparison of the temperature fields at $t = 10\Delta t = 0.5$  [s] with the initial temperature field $T(\bm x) = 0.5x^2 |(\sin(10x)+\cos(10y)|$ suggests that the fluctuation in the predictions concerning the reference solution by FEM is mitigated by enhancing training samples.
When we employed the homogeneous thermal conductivity, it was clear that the error sometimes decreased from the previous time steps. This can happen in the present framework since the input temperature fields at the subsequent time steps after the first step are likely to be similar to some of the training samples, resulting in a better interpolation accuracy in the network prediction than the previous step. We also give a detailed discussion on this point in the next subsection about the influence of time step size.
For the heterogeneous conductivity case, the prediction accuracy with the combination of Fourier and constant was low compared to the other three types of training samples. This is the same trend as for the homogeneous case. The contours on the right of the figure show that the NNs failed to correctly predict the temperature evolution, particularly in the low conductivity regions when only the Fourier and constant generators are used; see Fig.~\ref{err_samples_hetero}\,(a), which also means that the frequency in a variety of training samples greatly affects the prediction. Overall, the employment of diverse patterns for the training data sets was shown to be effective in improving the predictive performance of the present framework. 

\subsection{Influence of time step size}
\begin{figure}[H]
    \centering
    \includegraphics[width=140mm]{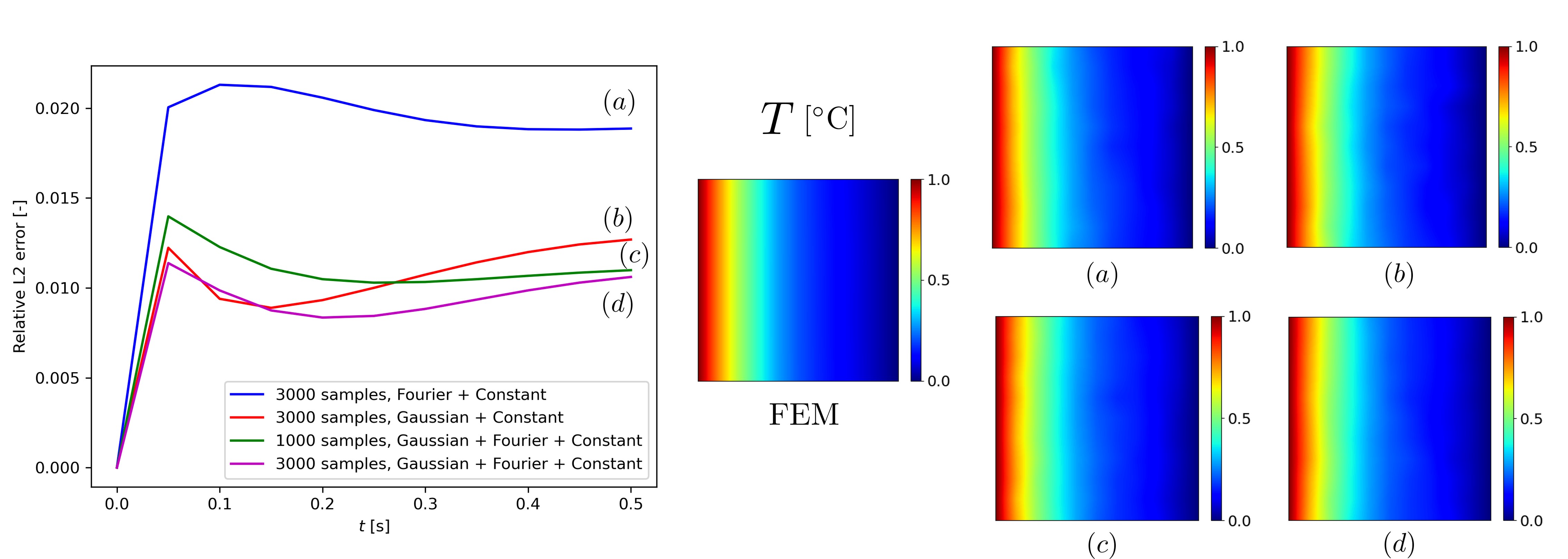}
    \caption{Left: Average relative L2 error norm from the five initial temperature fields over time in the case with the homogeneous thermal conductivity for different training data sets. Right: Temperature fields at $t = 10\Delta t = 0.5$  [s] when the initial temperature field $T(\bm x) = 0.5x^2 |(\sin(10x)+\cos(10y)|$ is given.}
    \label{err_samples_homoge}
\end{figure}

\begin{figure}[H]
    \centering
    \includegraphics[width=140mm]{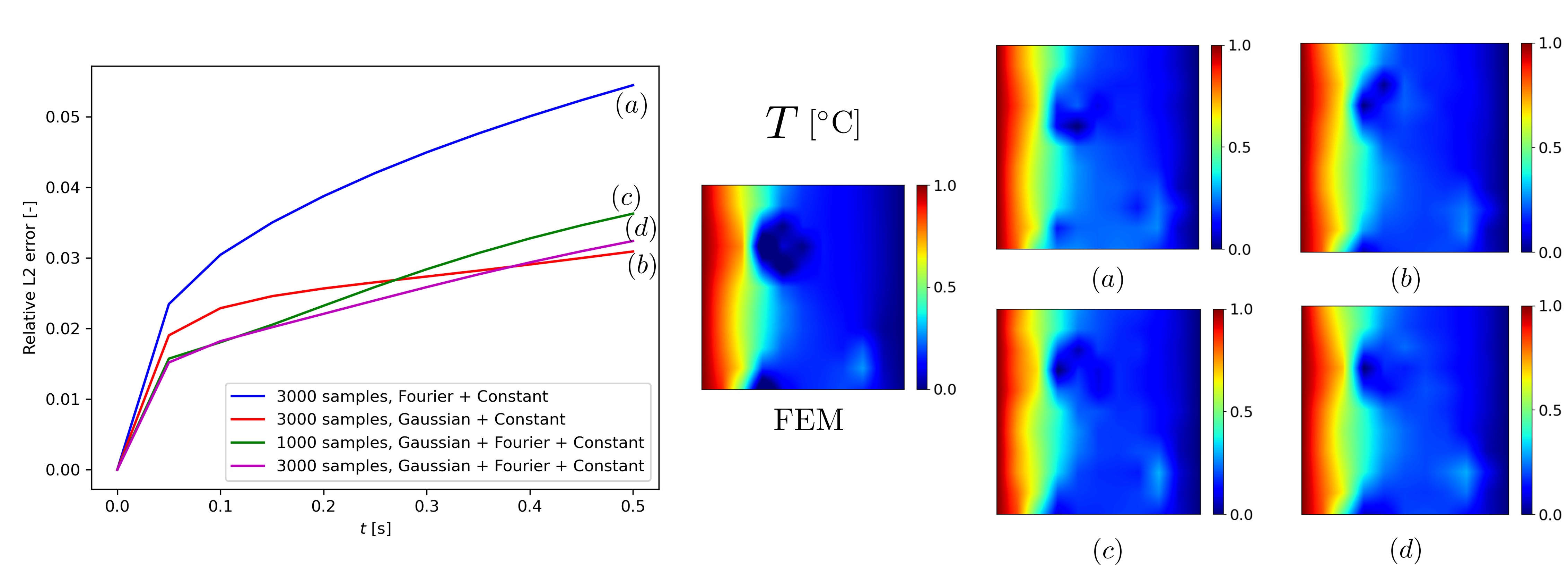}
    \caption{Left: Average relative L2 error norm from the five initial temperature fields over time in the case with the heterogeneous thermal conductivity for different training data sets. Right: Temperature fields at $t = 10\Delta t = 0.5$  [s] when the initial temperature field $T(\bm x) = 0.5x^2 |(\sin(10x)+\cos(10y)|$ is given.}
    \label{err_samples_hetero}
\end{figure}

\begin{figure}[H]
    \centering
    \includegraphics[width=140mm]{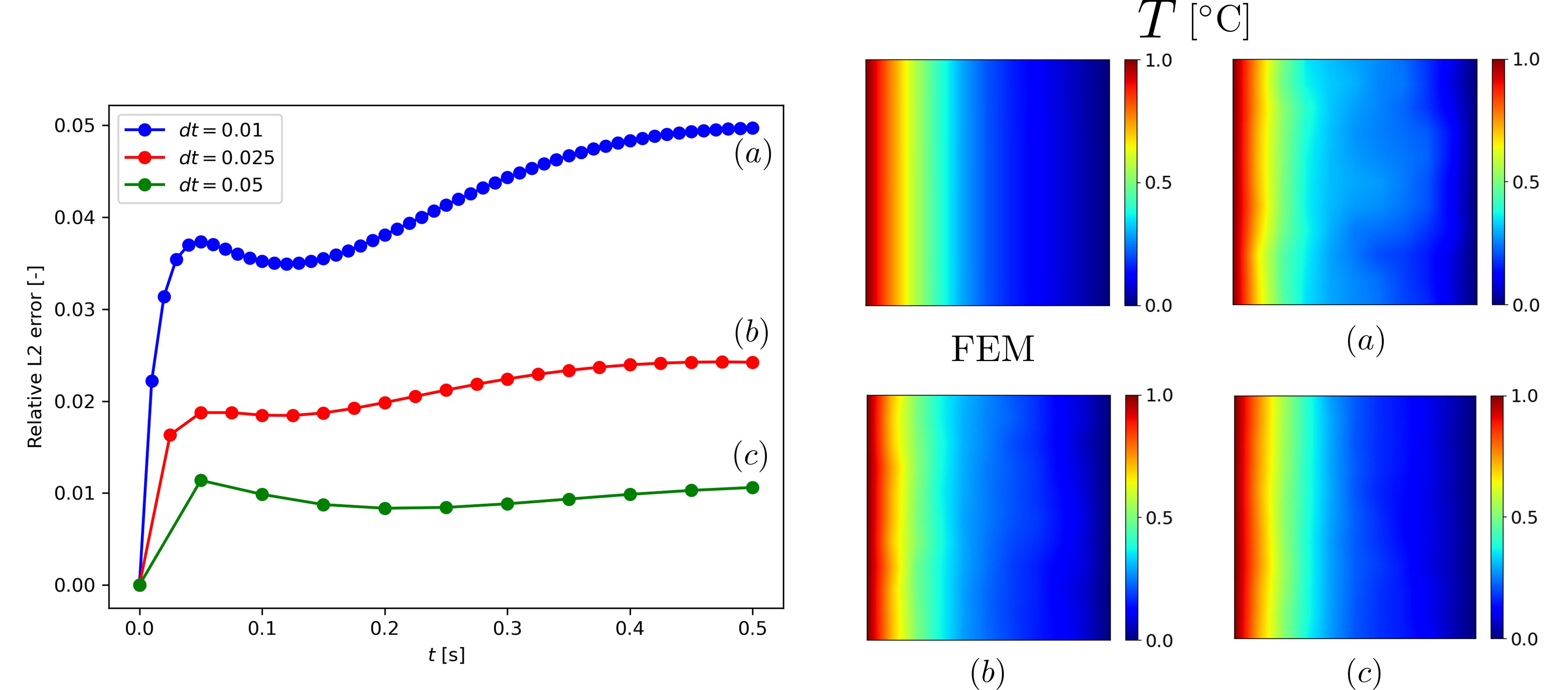}
    \caption{Left: Average relative L2 error norm from the five initial temperature fields over time in the case with the homogeneous thermal conductivity for three different time step sizes. Right: Temperature fields at $t = 10\Delta t = 0.5$  [s] when the initial temperature field $T(\bm x) = 0.5x^2 |(\sin(10x)+\cos(10y)|$ is given.  }
    \label{err_dt}
\end{figure}

The present FOL framework is flexible in terms of the time step size. Although one can choose an arbitrary time step size according to the time scale of the situation of interest, it is worth investigating how the choice of time step size influences the accuracy of the prediction over time, especially in terms of the number of time-marching steps. We used $3000$ training samples from the three temperature pattern generators, and the networks were trained for $1000$ epochs with homogeneous thermal conductivity.  The average relative L2 error over the results from the five initial temperature fields is shown on the right side of Fig. \ref{err_dt}, indicating that the error increases as the time step gets smaller. 
This can be partly attributed to the accumulation of errors due to successive inferences over time. In addition, the initial temperature field, being the most extreme of the input temperature fields presented to the framework throughout its temporal evolution, poses a significant challenge to the networks to accurately predict the subsequent state. This difficulty arises because the extreme temperature field lies within the sparse part of the training sample distribution. However, after the first time step, the error was reduced in each case, especially when $\Delta t = 0.01$, up to $t = 0.2$, which was also seen in Fig.~\ref{err_samples_homoge}. The latter can be explained by the distribution of training samples. As the temperature field approaches a steady state, the NNs are more likely to experience patterns similar to the input temperature field during the training phase. On the other hand, for $\Delta t = 0.01$, the error increases again after $t = 0.2$, which may be due to error accumulation from multiple inferences. The overall results suggest that the time step size needs to be adjusted according to the target phenomenon to decrease the error accumulation that affects the accuracy of the prediction.
In future studies, one could consider incorporating the time step size as an additional input and appropriately balancing the dynamical terms with the right-hand side of the equation using higher-order time integration algorithms.

\subsection{Influence of number of epochs and optimizer}
\begin{figure}[H]
    \centering
    \includegraphics[width=153mm]{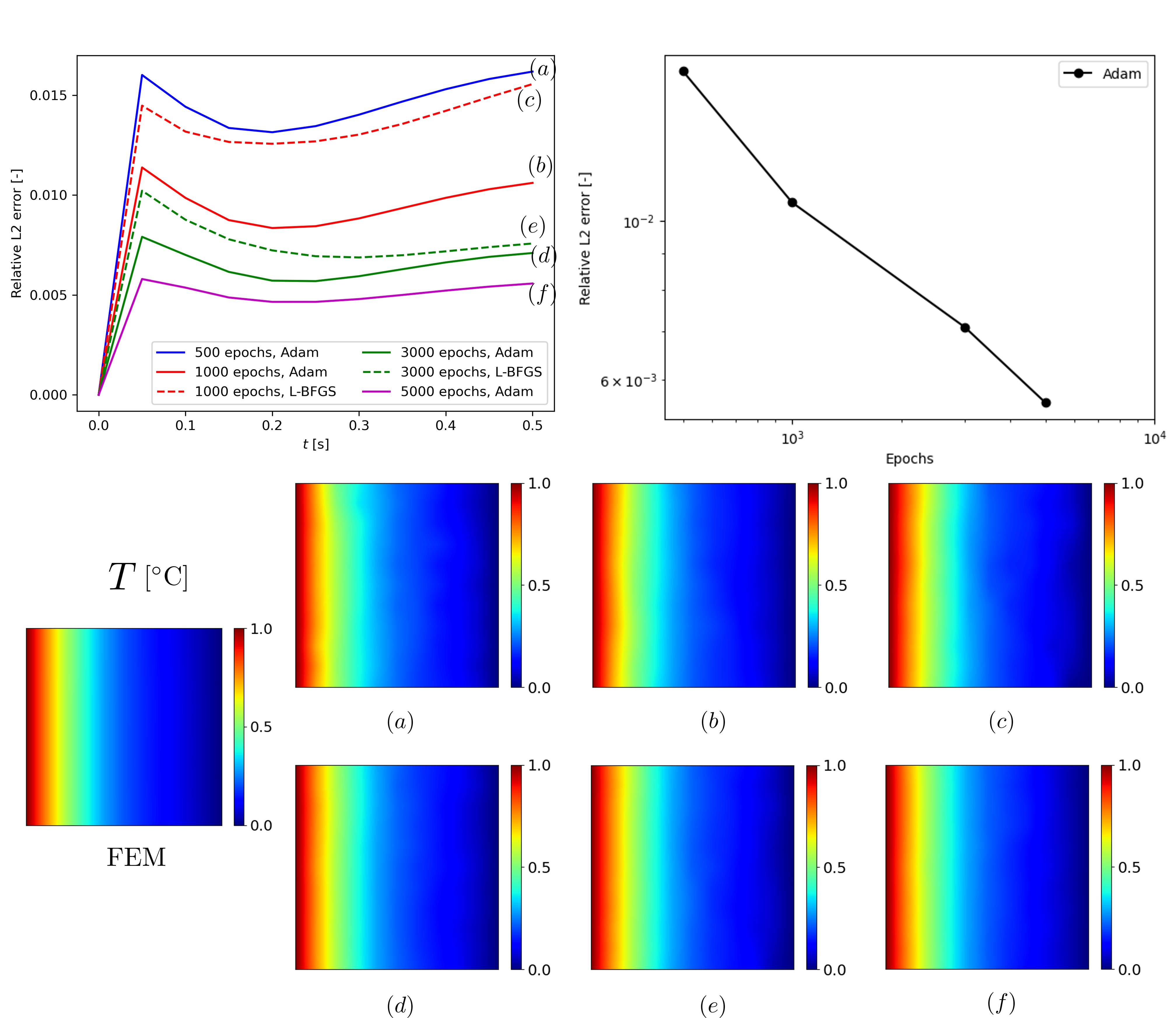}
    \caption{Left: Average relative L2 error norm from the five initial temperature fields over time in the case with the homogeneous thermal conductivity for different numbers of epochs and optimizers. Right: Relative L2 error with increasing number of epochs when Adam optimizer is employed. Bottom: Temperature fields at $t = 10\Delta t = 0.5$  [s] when the initial temperature field $T(\bm x) = 0.5x^2 |(\sin(10x)+\cos(10y)|$ is given. }
    \label{err_epochs_optimizer}
\end{figure}

The impact of the number of epochs and optimizers on prediction performance was also studied. The NNs were trained using $3000$ samples and a time step of $0.05$  [s]. The results for different numbers of epochs and two optimizers, Adam and L-BFGS (optimization algorithm employing quasi-Newton), are shown in Fig.~
\ref{err_epochs_optimizer}. The results indicate that increasing the number of epochs generally improved the predictive performance.  Looking into the detail shown in the log-log plot on the right of Fig.~\ref{err_epochs_optimizer} indicates the exponential decrease in relative L2 error with an increasing number of epochs. This implies that along with the loss history in Fig.~\ref{fig:loss_history}, even a slight decrease in the loss value, or namely the residual, improved predictive accuracy. One should also keep in mind that there is a trade-off relationship between the training cost and predictive accuracy. The temperature distributions on the bottom show that the main trend of the temperature evolution was sufficiently captured by FOL even with 500 and 1000 epochs, suggesting one does not need to train the NNs for too many epochs to have reasonable predictions. In terms of optimizers, a comparison between Adam and L-BFGS optimizers for the same number of epochs shows that Adam outperformed L-BFGS in FOL. This may be due to insufficient hyperparameter tuning for L-BFGS or the smoothness of the addressed optimization problem. Another reason for potential performance issues with L-BFGS optimization could be the approximation of the Hessian matrix, which estimates the curvature of the parameter space. As the parameter space increases in size, L-BFGS may not perform as well as ADAM.
A hybrid combination of Adam and L-BFGS can also be more promising in the future, see \cite{Harandi.2023, rathore2024challenges}.
\vspace{-4mm}

\subsection{Influence of the activation function}
\begin{figure}[H]
    \centering
    \includegraphics[width=160mm]{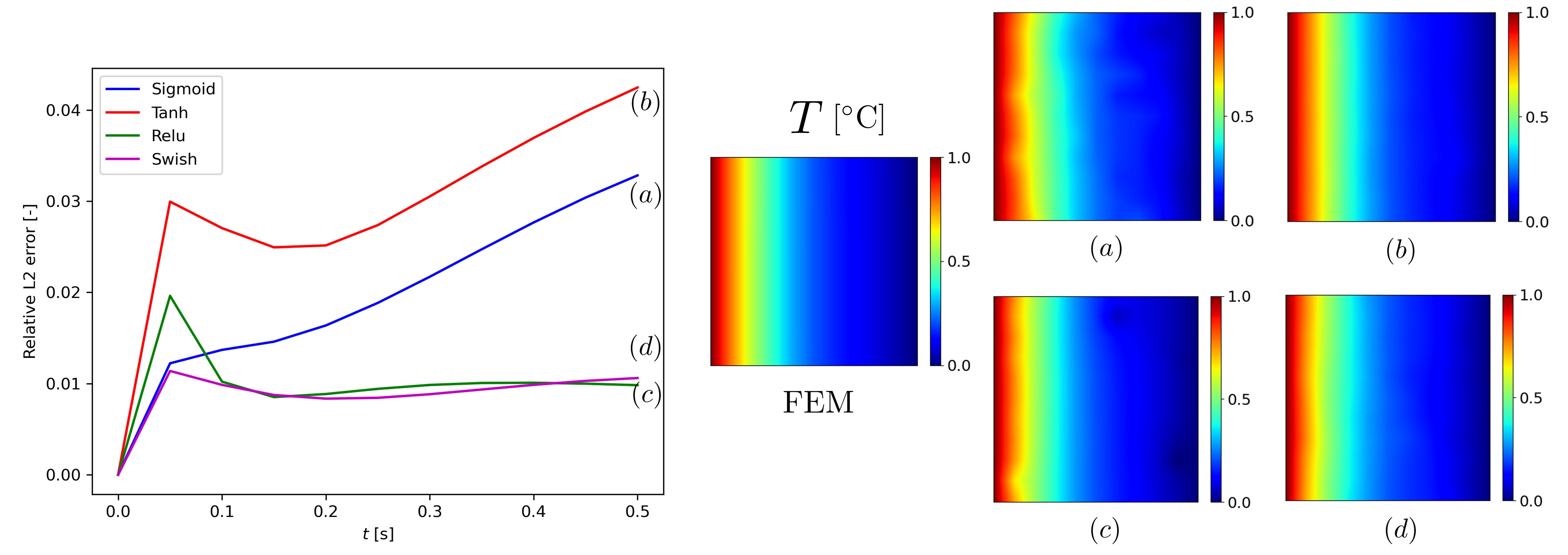}
    \caption{Left: Average relative L2 error norm from the five initial temperature fields over time in the case with the homogeneous thermal conductivity for three different activation functions. Right: Temperature fields at $t = 10\Delta t = 0.5$  [s] when the initial temperature field $T(\bm x) = 0.5x^2 |(\sin(10x)+\cos(10y)|$ is given. }
    \label{err_af}
\end{figure}

The study also investigated the impact of different activation functions. In addition to Swish, which was used in the other cases, the performance of sigmoid and hyperbolic tangent (tanh) functions were tested. The training was done for $1000$ epochs with $3000$ samples from the three types of temperature field generators. Fig. \ref{err_af} confirms that Swish outperformed sigmoid and tanh in terms of relative L2 error, averaged from the results of the five initial temperature cases. One possible reason for Swish's superior performance in this situation is the restricted temperature range between 0 and 1 in the present study. However, other activation functions may be viable options in problem setups with different value ranges. When comparing the performance of ReLU with Swish, it was observed that the error was noticeably larger with ReLU than with Swish at the first step. This could be due to the discontinuity at zero in the ReLU function, which affects learning in the vicinity of a temperature of 0 ° C.  

\subsection{Influence of network architecture}
\begin{figure}[H]
    \centering
    \includegraphics[width=160mm]{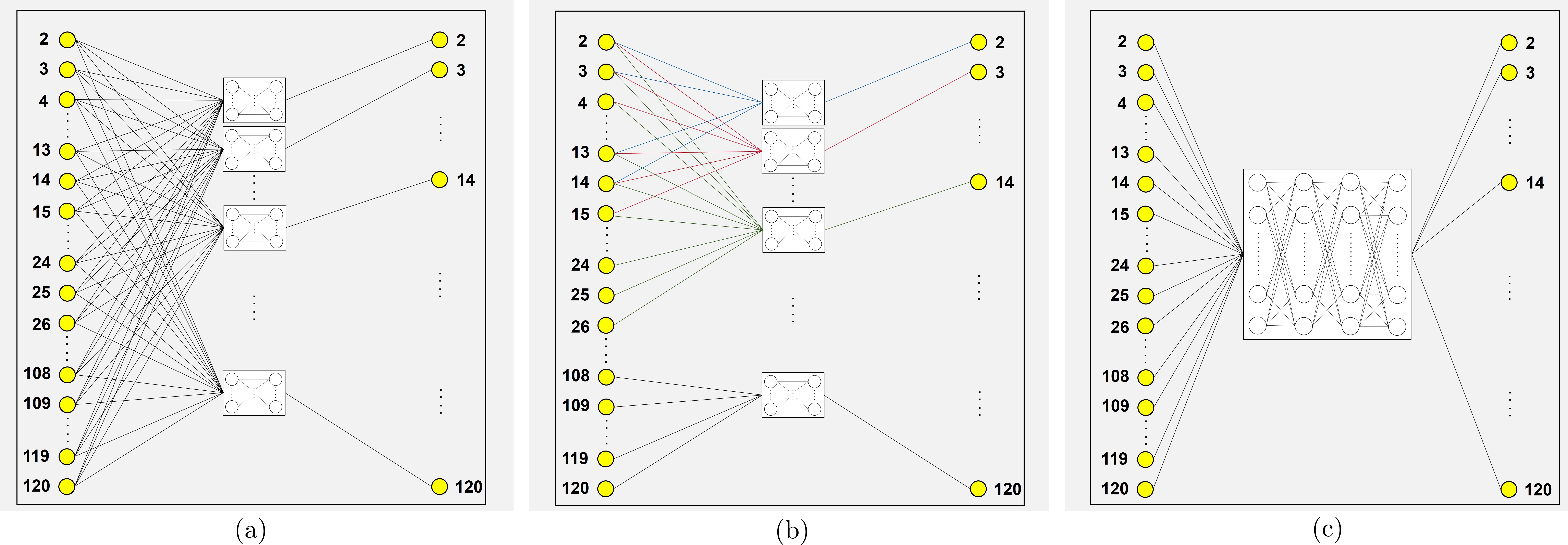}
    \caption{Comparison of the three network architectures: (a) original architecture, (b) elementwise-connected architecture, and (c) fully connected architecture. The input and output fields do not include Dirichlet boundaries due to the hard boundary condition.}
    \label{na_comparison}
\end{figure}

\begin{figure}[H]
    \centering
    \includegraphics[width=140mm]{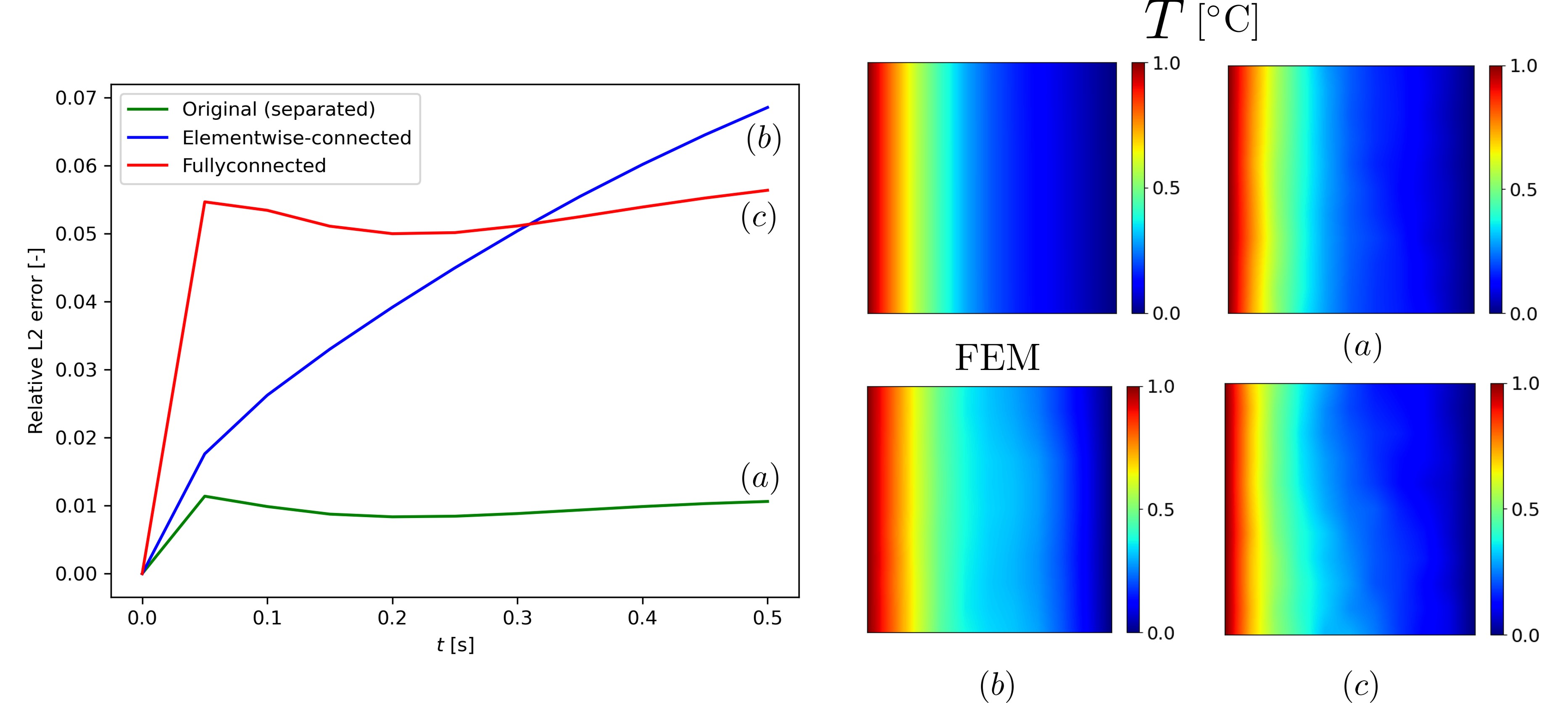}
    \caption{Left: Average relative L2 error norm from the five initial temperature fields over time in the case with the homogeneous thermal conductivity for three different network architectures. Right: Temperature fields at $t = 10\Delta t = 0.5$ [s] when the initial temperature field $T(\bm x) = 0.5x^2 |(\sin(10x)+\cos(10y)|$ is given.}
    \label{err_na}
\end{figure}

The present framework consists of separate NNs for corresponding nodal temperatures at the next time step. On this matter, other network architecture options could be conceived, such as a fully connected one that returns the whole output field from a single NN. To quantitatively ensure the superiority of the present NN architecture, we evaluated the three types of network architectures, including the original one used for the main study \cite{rezaei2024integration}. The created network architectures are compared in Fig.~\ref{na_comparison}. Compared to the original architecture, the elementwise-connected architecture takes nodal temperatures from adjacent elements as input for the output nodal temperature. For the fully connected architecture, nodal temperatures are provided as inputs and the entire nodal temperature field for the next time step is returned as output. 
In this study, 4 layers with 170 neurons in each layer were selected to ensure that the number of trainable parameters is comparable with each other. As a result, the fully connected architecture has 121,139 trainable parameters, whereas the separated and elementwise-connected architectures have 110,979 trainable parameters.
It is noted that the nodal number starts with 2 as we apply hard constraints for the Dirichlet boundary conditions and, therefore, eliminate those nodes from the training targets, as explained in Section 3.2.
The results shown in Fig. \ref{err_na} indicate that the two network architectures newly introduced here led to worse accuracy compared to the original architecture approximately by a factor of 5. Separating NNs for different outputs while considering the entire problem domain for input is the most effective way to learn the mapping between the input and output physical fields in FOL. 
In terms of the training cost in the above cases, on the other hand, the fully connected architecture spent 6 hours 23 minutes 54 seconds, whereas the separated architecture took 7 hours 29 minutes 30 seconds and the elementwise connected architecture 7 hours 40 minutes 32 seconds when they were trained with SciANN for 1000 epochs on a single GPU node of NVIDIA GeForce RTX 2080 12GB, indicating that the fully connected one can be an efficient option that can still serve as a surrogate with sufficiently high-accuracy prediction. This is also supported by the comparison of the inference costs, where the fully connected architecture was shown to be 3.103 times faster than the separated network architecture in the measurement. Regarding the improvement of the separated and elementwise-
\noindent
connected architectures, one interesting direction in the future would be to investigate whether increasing the size of the element groups in Fig.~\ref{na_comparison} (b) can enhance predictive accuracy, which is in the end identical to that of Fig.~\ref{na_comparison} (a).
Reducing the number of input dimensions to each NN shown in Fig.~\ref{na_comparison} (a) and (b) leads to the reduction of parameters in NNs in the present framework, resulting in less training cost.
This improvement would play a role when one wants to apply FOL to a model with more nodes than what is considered in this work. Furthermore, it would also be beneficial to even reduce the number of neurons or layers in each network architecture, leading to the prevention of overfitting in NNs.
Nevertheless, for large-scale problems in which a large number of degrees of freedom need to be taken into account, one would have more benefits with the fully connected architecture, especially in combination with learning in latent space by employing techniques to condense information such as autoencoder.

\subsection{Influence of mesh size}
\begin{figure}[H]
    \centering
    \includegraphics[width=160mm]{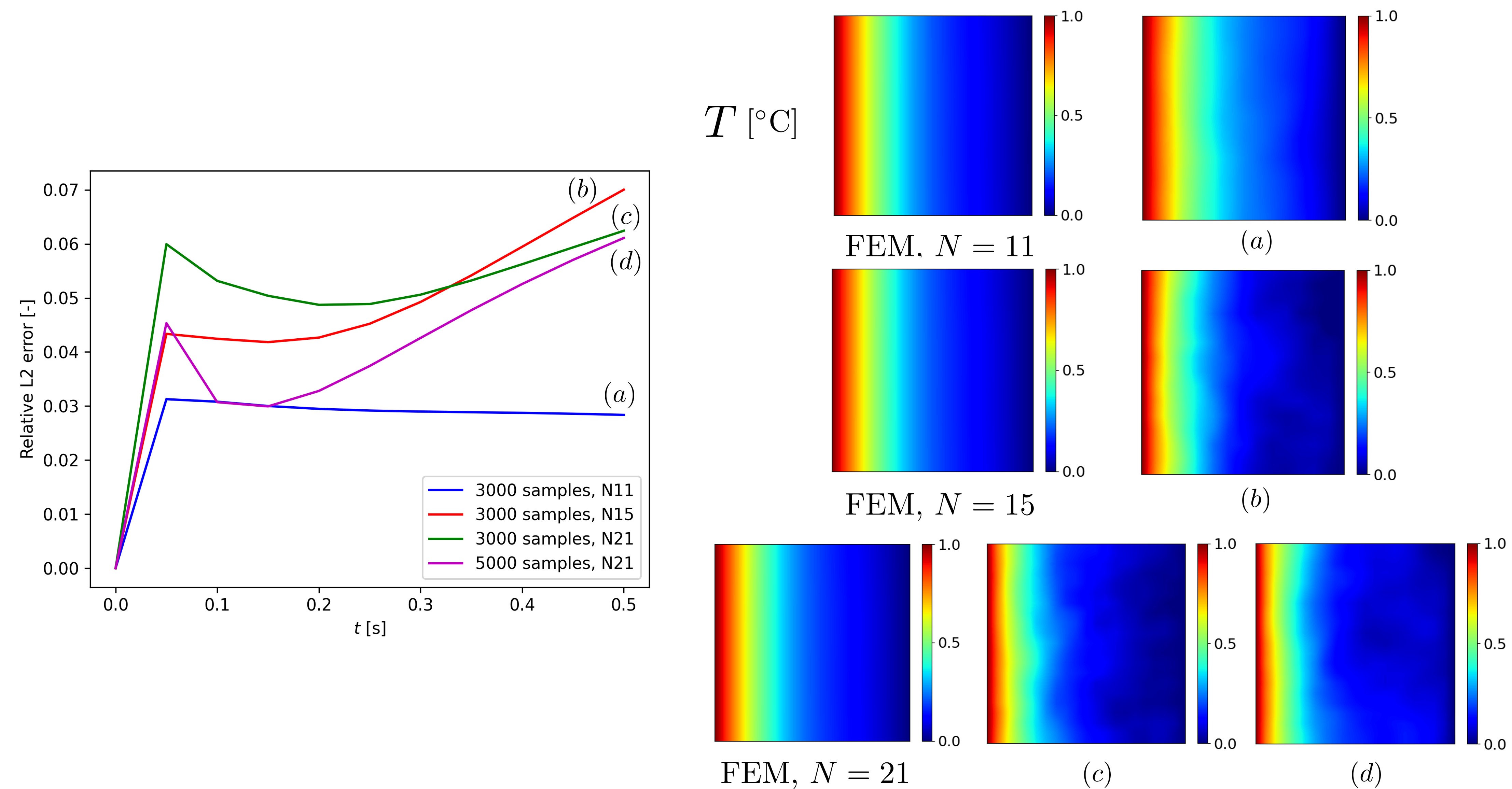}
    \caption{Left: Average relative L2 error norm from the five initial temperature fields over time in the case with the homogeneous thermal conductivity for three different mesh sizes. Right: Temperature fields at $t = 10\Delta t = 0.5$ [s] when the initial temperature field $T(\bm x) = 0.5x^2 |(\sin(10x)+\cos(10y)|$ is given. }
    \label{err_ms_homoge}
\end{figure}

\begin{figure}[H]
    \centering
    \includegraphics[width=160mm]{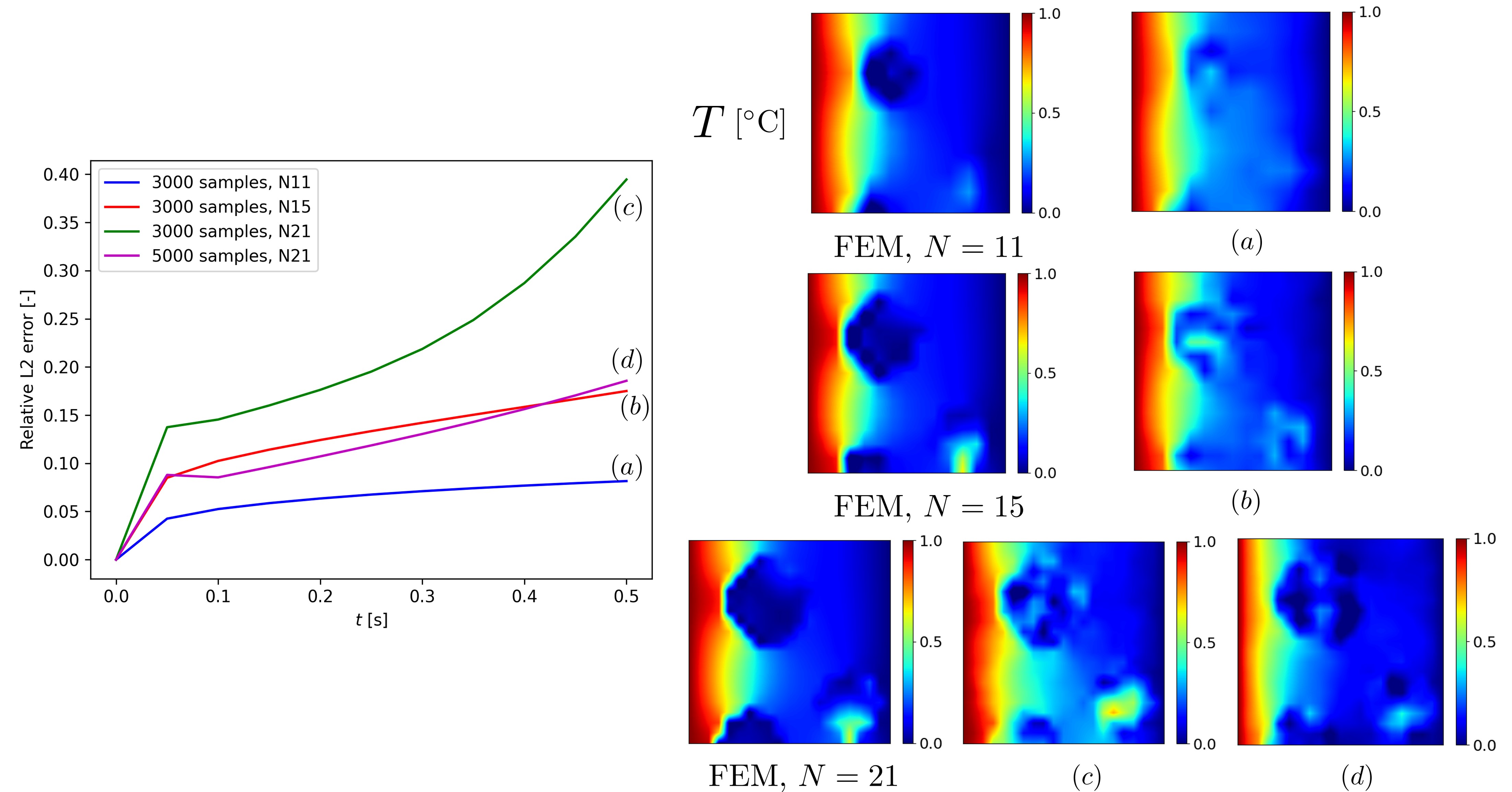}
    \caption{Left: Average relative L2 error norm from the five initial temperature fields over time in the case with the heterogeneous thermal conductivity for three different mesh sizes. Right: Temperature fields at $t = 10\Delta t = 0.5$ [s] when the initial temperature field $T(\bm x) = 0.5x^2 |(\sin(10x)+\cos(10y)|$ is given. }
    \label{err_ms_hetero}
\end{figure}

\begin{figure}[H]
  \begin{minipage}{0.49\textwidth}
    \centering
    \includegraphics[width=74mm]{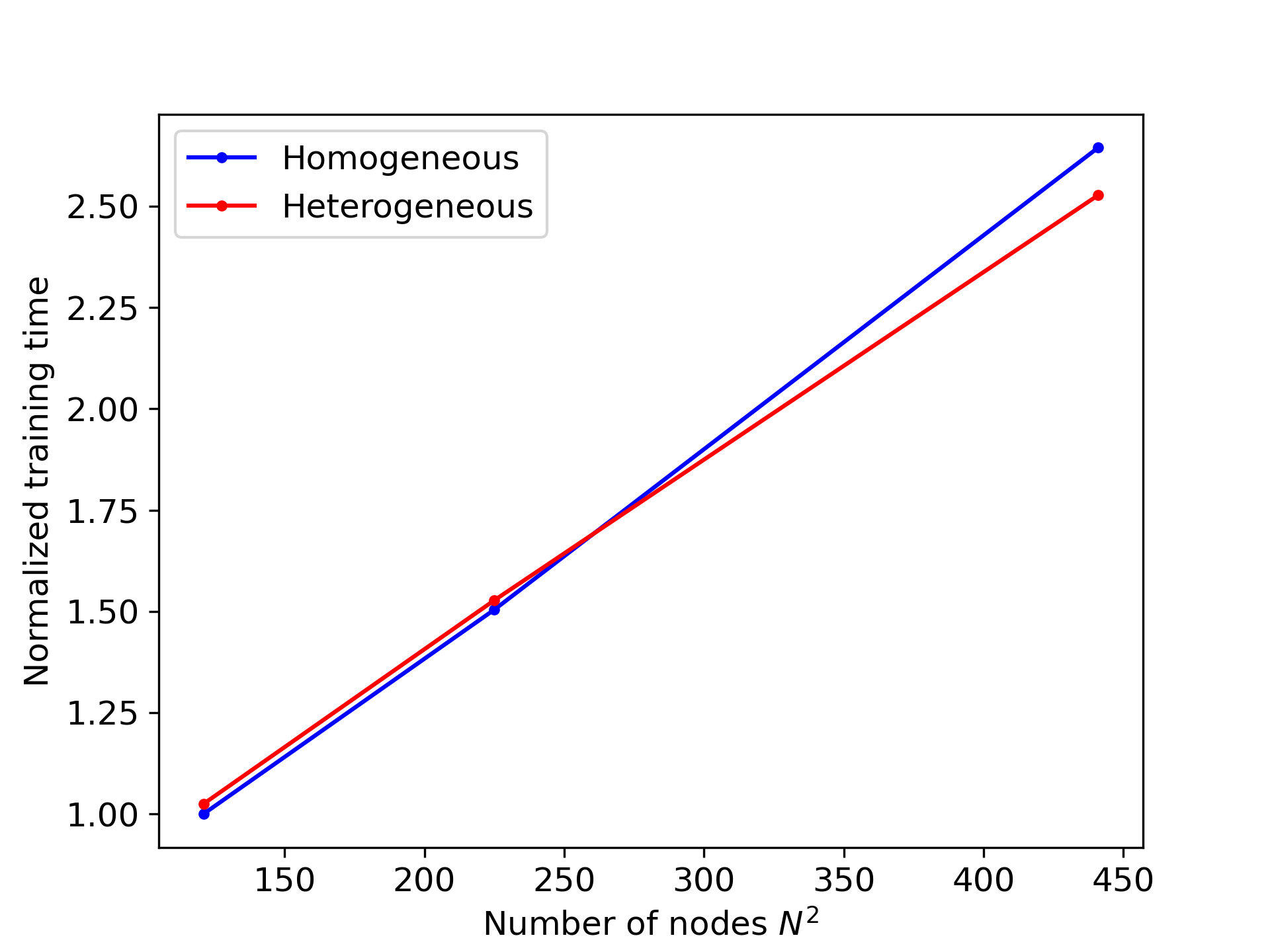}
    \caption{Normalized training time for three different mesh sizes with 3000 samples and a batch size of 60.}
    \label{training_time}
  \end{minipage}
  \hfill
  \begin{minipage}{0.49\textwidth}
    \centering
    \includegraphics[width=74mm]{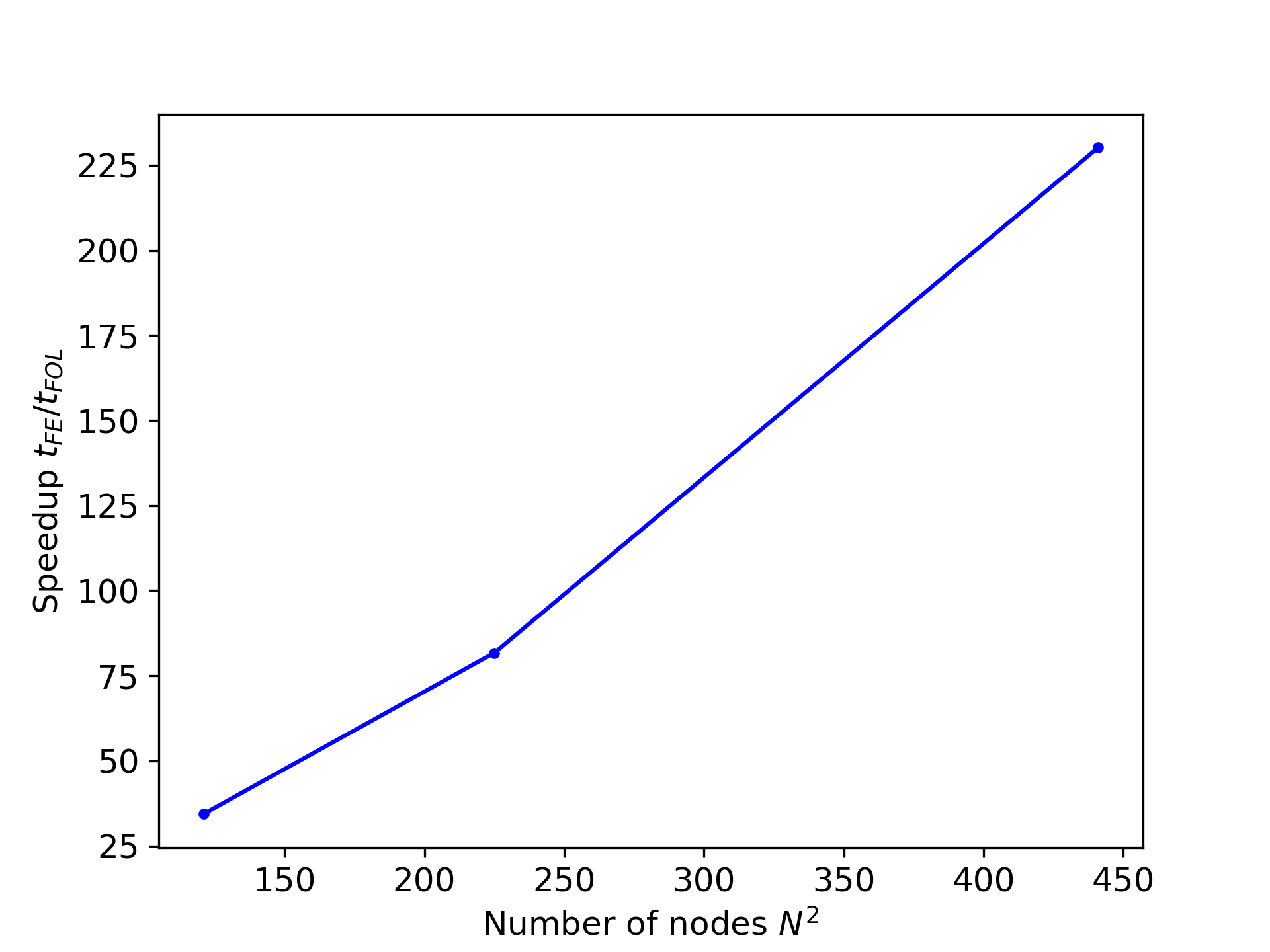}
    \caption{Speedup of the FOL evaluation time compared to the FE calculation time for three different mesh sizes.}
    \label{evaluation_speedup}
  \end{minipage}
\end{figure}

Since the mesh size affects the prediction accuracy and the subsequent training cost, it is also pivotal to gain insights on this point. 
For that, three different element sizes, one is the same one as in the main study ($N = 11$) and the others are finer than the original one ($N = 15, 21$), on the squared domain were considered to perform the comparison.
Due to the increase in the training cost for finer mesh, we performed this study on the JAX platform which has the equivalent architecture to the SciANN-based code for faster training. 
The fully connected network architecture with the number of neurons in each layer set to 170 and 4 layers was employed for the study.
The batch size was set to 60 for 3000 samples and 100 for 5000 samples to be consistent in terms of the batch size ratio to the total number of samples. 
The obtained average relative L2 errors along with the temperature distribution at $t = 10 \Delta = 0.5 $ s for the initial temperature field $T(\bm x) = 0.5x^2 |(\sin(10x)+\cos(10y)|$ are shown in Figs. \ref{err_ms_homoge} and \ref{err_ms_hetero} for the homogeneous and heterogeneous thermal conductivities, respectively.
In Fig. \ref{err_ms_homoge}, the magnitudes of the average relative L2 error norm over time exhibited noticeable differences between $N = 11$, $N = 15$, and $N = 21$. This could be due to the reduced number of trainable parameters in each nodal evaluation for finer meshes. One can also confirm that by enhancing training samples in the case of $N = 21$ a decent improvement in overall prediction accuracy is achieved.
On the other hand, in Fig. \ref{err_ms_hetero} the error started to blow up after several time steps when 3000 training samples were utilized for training in the case of $N = 21$. This was significantly mitigated by enhancing the training samples from 3000 to 5000, part of which is through adding higher frequencies to the Fourier series sample generator.
This indicates that in the heterogeneous case, the quality of training samples critically affects the prediction accuracy in the finer mesh as shown in the case of $N = 21$, which is different from the observation in Fig. \ref{err_samples_hetero} where one does not see such an extreme error evolution. 
When it comes to the training cost, one has to pay the price for increasing the number of nodes in FOL. As shown in Fig.~\ref{training_time}, the training time increased linearly with increasing number of nodes. The homogeneous case with $N=21$ required 2.645 times more training time than the same setup with $N=11$. The same trend was confirmed for the heterogeneous case.
It is noted that the training was done within 30 minutes for all four cases using JAX.
On the other hand, the more the number of nodes increases, the more speedup FOL can achieve against FEM as shown in Fig.~\ref{evaluation_speedup}, given the same network architecture with varying input and output dimensions. This indicates the strong potential of FOL as a fast surrogate to conventional FE solvers for large models with a large number of nodes.
In application scenarios, one has to therefore adapt the variety of training samples to achieve reasonable prediction, especially when considering heterogeneous domains. Further exploring the integration of this method with data-driven auto-encoders presents intriguing possibilities to address potential constraints associated with increased mesh densities (i.e. higher resolutions), as highlighted in \cite{liu2024multi, kontolati2024learning,koopas2024introducing}.



\subsection{Capability of handling arbitrary domains}
\begin{figure}[H]
    \centering
    \includegraphics[width=160mm]{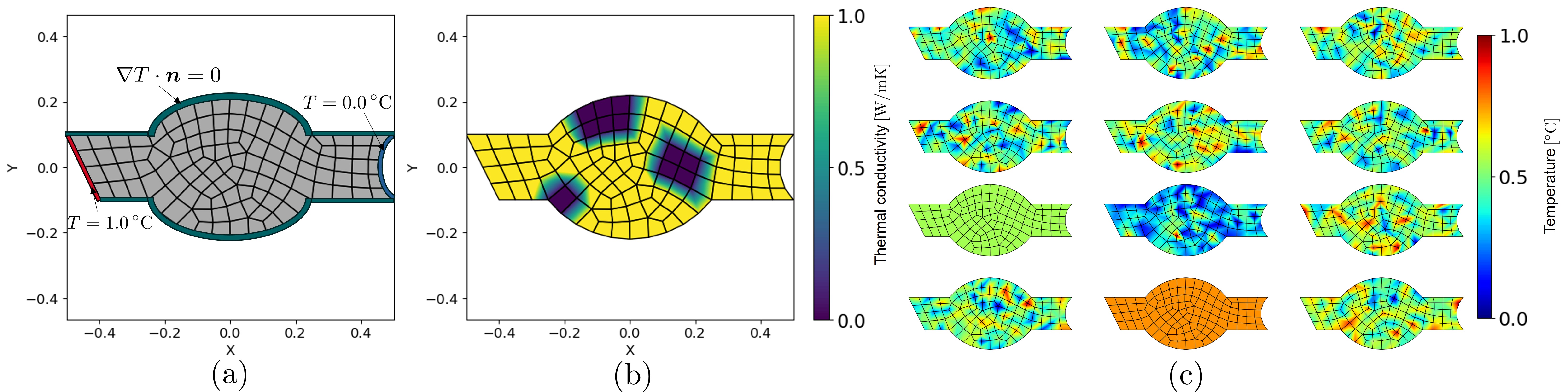}
    \caption{(a) Irregular domain discretized by quadrilateral elements and prescribed boundary conditions. (b) Introduced heterogeneous thermal conductivity. (c) Examples of the training samples for the irregular domain.}
    \label{arbi_setup_samples}
\end{figure}

One of the main advantages of leveraging FEM in the context of finite-dimensional operator learning lies in the capability of handling arbitrary domains. To demonstrate the applicability of the present framework to such a scenario, we considered a different domain, as shown in Fig. \ref{arbi_setup_samples} (a). In addition, we introduced heterogeneity of thermal conductivity, shown in Fig. \ref{arbi_setup_samples} (b), to this problem setup. The training was first performed with 3000 samples and $\Delta t = 0.05$ [s] for 1000 epochs. Here, we generated the training samples from Gaussian and constant-temperature generators, not using the Fourier series, to reduce the complexity of sample generation, some of which are shown in Fig. \ref{arbi_setup_samples} (c). Additionally, please refer to the discussions in Section 4.1, where we demonstrated that accurate predictions can be obtained without relying on Fourier series-based samples. We tested the performance of the networks for the initial temperatures of $T(\bm x) = 0.5$ and $T(\bm x) = |\sin(10 x)|$. We confirmed that in both cases, as 
\begin{figure}[H]
    \centering
    \includegraphics[width=160mm]{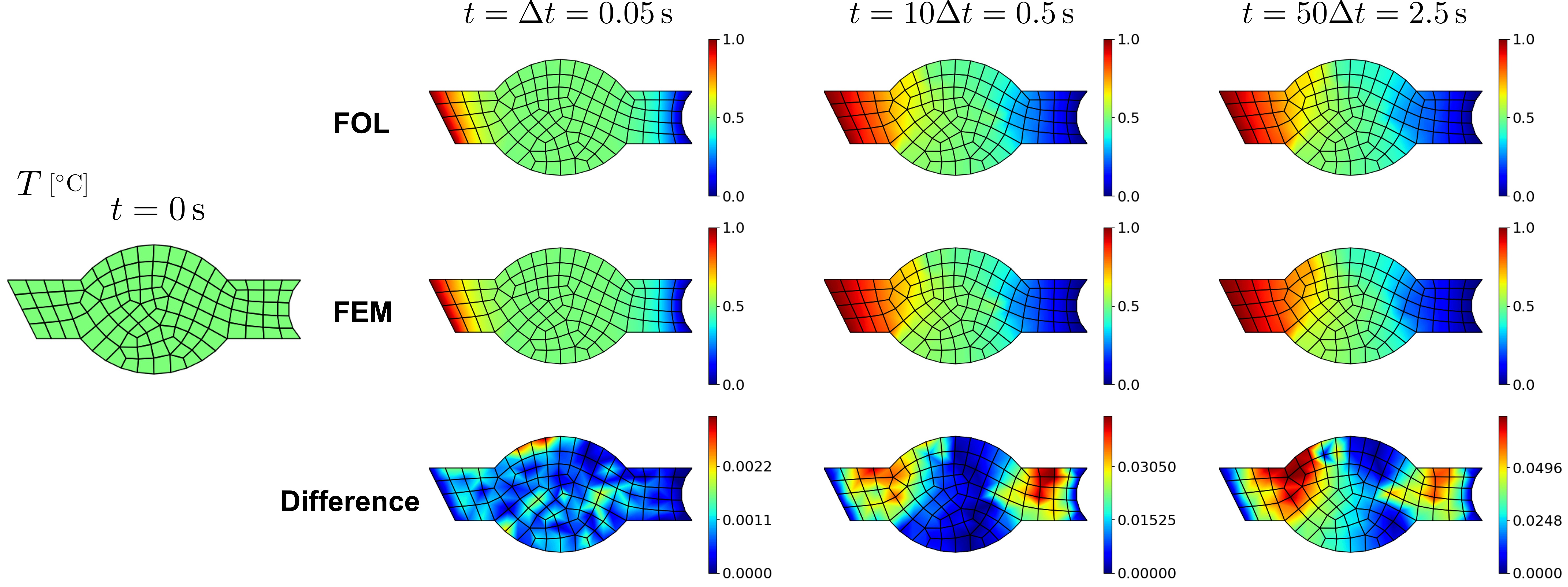}
    \caption{Temperature evolution from the FOL prediction (top),  FE solution (middle), and the difference with $T(\bm x) = 0.5$ as an initial temperature field.}
    \label{arbi_const_temp}
\end{figure}

\begin{figure}[H]
    \centering
    \includegraphics[width=160mm]{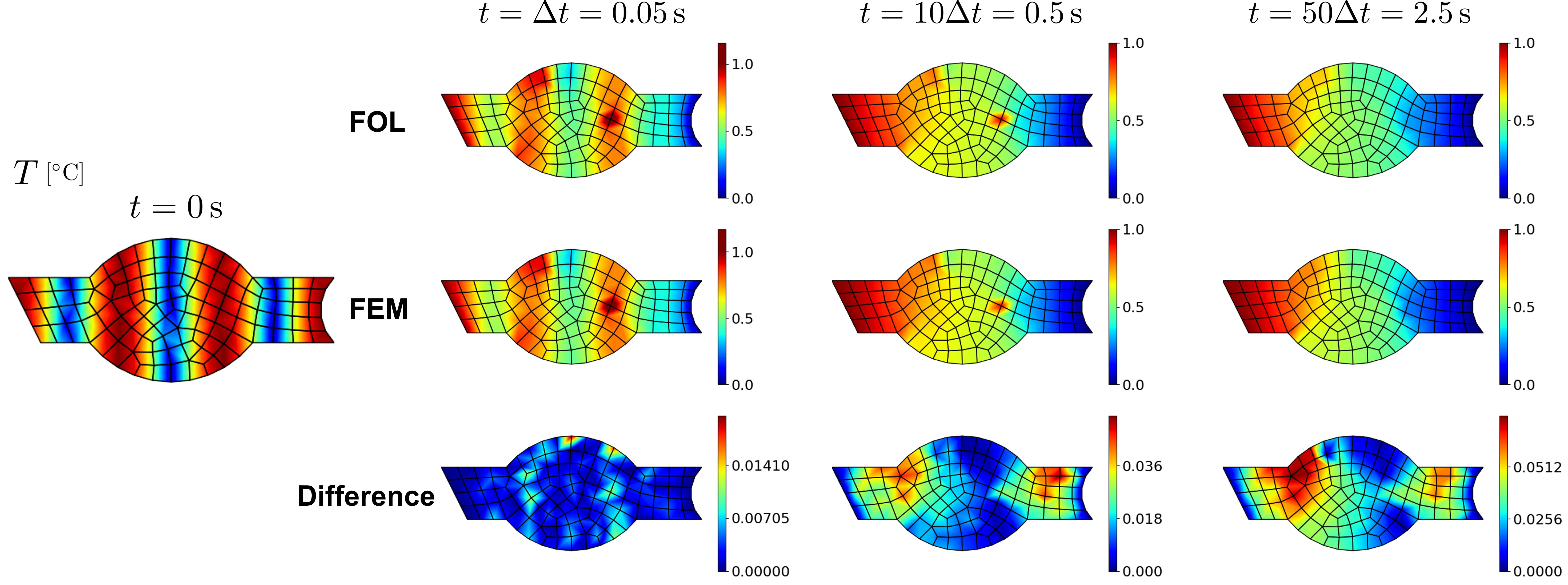}
    \caption{Temperature evolution obtained from the FOL prediction (top),  FE solution (middle), and the difference with $T(\bm x) = |\sin(10 x)|$ as an initial temperature field.}
    \label{arbi_sin_temp}
\end{figure}
\noindent
shown in Figs. \ref{arbi_const_temp} and \ref{arbi_sin_temp}, the overall solution trend obtained by the proposed FOL framework agreed with the solution by FEM with the maximum absolute error around 0.1. The error was concentrated in the upper right and left part of the domain at $t = 10 \Delta t = 0.5$ and $t = 50 \Delta t = 2.5$ [s], which is also explained by the steep change in the temperature evolution due to the presence of heterogeneity.
Furthermore, the error magnitudes did not differ very much from the results with the square domain as in Figs. \ref{hetero_sin}, \ref{hetero_gauss}, and \ref{hetero_trig}. The results of Fig. \ref{arbi_sin_temp} are particularly noteworthy. This proves that FOL works for a problem setup with complex initial temperature, heterogeneity, and irregular domain discretized by unstructured mesh without cumbersome modification to the framework.

\subsection{Computational cost and advantages of proposed framework}
The main goal of this work is to establish a surrogate model for conventional numerical solvers. In this context, the prediction by the networks should be faster than the solution by numerical analysis. To quantitatively evaluate the speed of obtaining solutions by the FOL framework, the runtimes of the network inference with the same network architecture and finite element calculation were measured by performing the same task. The measurement was performed on the same CPU platform and environment to ensure the fairness of the measurement. We assumed ten-time steps in solving the heat equation with FEM. Therefore, the same network evaluation was performed ten times as well. As a result, the prediction time with the separated network architecture was 10.8 times faster than FEM for the same setup. This result suggests that the network has the potential to be used as a surrogate for classic numerical solvers. 

Although a faster inference than solving with a classical solver can be achieved with the proposed framework, one has to train the NNs for a relatively long time. For example, the training time for 1000 epochs on the problem setup shown in Figs.~\ref{dimensions_bc} and \ref{fig:grid} took approximately 7 hours on a single GPU node of NVIDIA GeForce RTX 2080 12GB. However, the training time will be much faster using JAX (high-performance machine learning framework) which has superior features for deep learning, such as just-in-time compilation and vectorization. 
In addition, training NNs in FOL is usually a one-time investment. Once trained, users can use it for any input and can obtain the solution much quicker than numerical solvers, even for a model that requires many nodes to solve accurately.

As a result, the developed physics-informed operator learning framework has several advantages over other deep learning-based methods. First, the training of the networks is completely unsupervised. Unlike data-driven deep learning models, there is no need to prepare an extensive dataset from costly simulations or experiments. Instead, a dataset of random temperature patterns generated by the Gaussian random process and Fourier series combined with constant temperature fields is used for training. This approach allows for covering a wide range of possible temperature cases without relying on labeled data. Additionally, the framework utilizes shape functions for spatial discretization and backward difference approximation for temporal discretization. The resulting pure algebraic equation, similar to data-driven loss functions, eliminates the need for time-consuming automatic differentiation during weight and bias optimization, resulting in faster training. Furthermore, as shown in the previous subsection, the present framework can handle irregular domains quite easily, along with heterogeneity in the domains, thanks to the feature of the finite element method, which will be helpful in many engineering applications. Lastly, other types of spatiotemporal PDEs, such as the Allen-Cahn equation or Cahn-Hilliard equation, could be incorporated into this framework, given corresponding finite element formulations. This makes the proposed framework usable in the context of other physics. 

\section{Conclusion}
This study has presented a novel physics-informed operator learning framework based on the finite element discretization scheme for spatiotemporal PDEs. After training with various temperature fields, including those generated by Gaussian distribution and Fourier series, as well as constant temperature fields, the network can accurately predict dynamic temperature evolutions for any arbitrary temperature input within the assumed temperature range. This is achieved with a relative L2 error below 0.1 in most cases, without the need for retraining under fixed boundary conditions and domain. The applicability of the method to heterogeneous heat conductivity and irregular domains is also confirmed. Additionally, the suggested network design can achieve over ten times the speed of the corresponding FEM solver on the same platform. It is important to note that the training is conducted entirely without ground truth data obtained from numerical simulations, making the framework a completely unsupervised learning approach. Furthermore, the training efficiency is enhanced compared to other operator learning approaches that rely on time-consuming automatic differentiation. This is because the current framework uses FE-based discretization for space and backward difference approximation for time.
To summarize, this work explores the development of deep learning-based surrogates for dynamic physical phenomena without the need for supervised learning.

\begin{figure}[t]
    \centering
    \includegraphics[width=160mm]{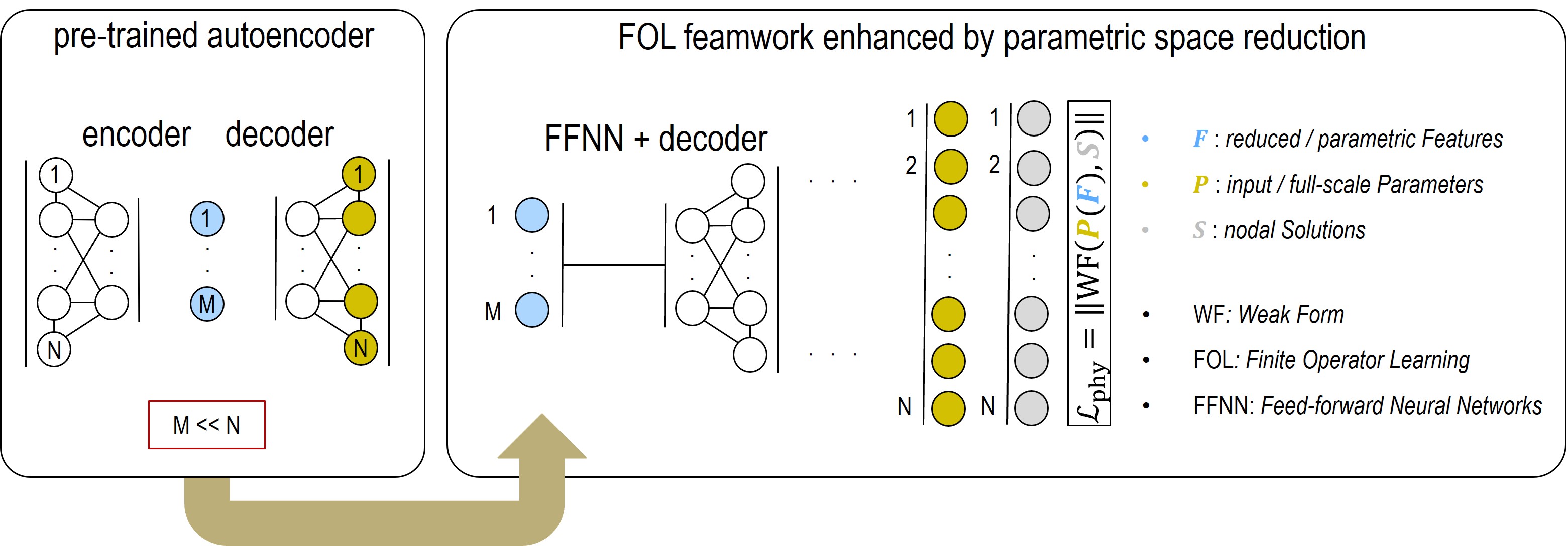}
    \caption{Schematic of employing autoencoder in FOL for efficient learning in a reduced parametric space.}
    \label{fig:outlook_autoencoder}
\end{figure}

On the other hand, although the proposed framework offers useful features, there are still some limitations that could be addressed in future work. Firstly, heat conductivity can also be a target of training in addition to the temperature field. This makes the framework flexible for various micro morphologies with phase-field modeling in mind. It may be possible to improve accuracy by implementing a higher-order temporal discretization scheme. 
One could also think about different network architectures that take multiple temperature fields as input.
Moreover, this study focused on transient heat conduction to showcase the performance of the framework. The present framework could be extended to other types of spatiotemporal PDEs, such as the convection-diffusion equation, the Allen-Cahn equation, or the Cahn-Hilliard equation, as the framework is developed with the aim of a generic deep learning framework for spatiotemporal dynamics phenomena described by PDEs. 
Additionally, although the present study only utilizes the bilinear interpolation that works for the majority of the possible applications, this framework could also be combined with higher-order basis functions such as the quadratic one for a better representation of geometry with curvature and further accuracy in prediction.
Lastly, to efficiently handle large models with a large number of nodes, a reduced parametric space illustrated in Fig.~\ref{fig:outlook_autoencoder} could be introduced by employing techniques such as autoencoder \cite{kontolati2024learning, koopas2024introducing, rezaei2024integration_mech}. 
\\\\
\textbf{Author Statement}:
A.H. and Sh.R. conceptualized the study. Y.Y., A.H., and Sh.R. developed the methodology. Y.Y. and Sh.R. developed the software.  M.M. provided the computational resources for conducting the study. Y.Y. conducted the formal analysis and investigation.  Y.Y., A.H., and Sh.R. wrote the original draft.  All authors reviewed and edited the manuscript. Sh.R. supervised the project.
\\ \\ 
\textbf{Competing Interests}:
The authors declare no competing financial or non-financial interests.
\\ \\
\textbf{Data Availability}:
The codes and data associated with this research are available upon request and will be published online following the official publication of the work.
\\ \\
\textbf{Acknowledgements}:
The authors would like to thank the Deutsche Forschungsgemeinschaft (DFG) for the funding support provided to develop the present work in the project Cluster of Excellence “Internet of Production” (project: 390621612). The authors also acknowledge the financial support of Transregional Collaborative Research Center SFB/TRR 339 with project number 453596084 funded by DFG gratefully. Mayu Muramatsu acknowledges the financial support from the JSPS KAKENHI Grant (22H01367). 

\bibliographystyle{elsarticle-num} 
\bibliography{bibtex/reference.bib}





\end{document}